\documentclass{article}

\PassOptionsToPackage{numbers,compress}{natbib}

\usepackage[preprint]{neurips_2021}

\usepackage[utf8]{inputenc} 
\usepackage[T1]{fontenc}    
\usepackage{hyperref}       
\usepackage{url}            
\usepackage{booktabs}       
\usepackage{amsfonts}       
\usepackage{nicefrac}       
\usepackage{microtype}      
\usepackage{xcolor}         

\usepackage{graphicx} 
\usepackage{caption,subcaption}
\usepackage{multirow}
\usepackage{graphicx}
\usepackage{multicol}
\usepackage{amsmath}
\usepackage{wrapfig}
\DeclareMathOperator*{\argmax}{arg\,max}

\makeatletter
\newcommand*{\rom}[1]{\expandafter\@slowromancap\romannumeral #1@}
\makeatother

\newtheorem{proof}{Proof}

\title{Rethink, Revisit, Revise: A Spiral Reinforced Self-Revised Network for Zero-Shot Learning}

\author{Zhe Liu, Yun Li, Lina Yao, Julian McAuley, Sam Dixon
}

\begin{document}

\maketitle

\begin{abstract}
Current approaches to Zero-Shot Learning (ZSL) struggle to learn
generalizable semantic knowledge capable of capturing
complex correlations. Inspired by \emph{Spiral Curriculum}, which enhances learning processes by revisiting knowledge, we propose a form of spiral learning which revisits visual representations based on a sequence of attribute groups (e.g., a combined group of \emph{color} and \emph{shape}). Spiral learning aims to learn generalized local correlations, enabling models to gradually enhance global learning and thus understand complex correlations. Our implementation is based on a 2-stage \emph{Reinforced Self-Revised (RSR)} framework: \emph{preview} and \emph{review}. RSR first previews visual information to construct
diverse attribute groups in a weakly-supervised manner. Then, it
spirally learns refined localities based on attribute groups and uses localities to revise global semantic correlations. Our framework outperforms state-of-the-art algorithms on four benchmark datasets in both zero-shot and generalized zero-shot settings, which demonstrates the effectiveness of spiral learning in learning generalizable and complex correlations. We also conduct extensive analysis to show that attribute groups and reinforced decision processes can capture complementary semantic information to improve predictions and aid explainability.
\end{abstract}

\section{Introduction}


Zero-Shot Learning (ZSL) aims to learn general semantic correlations between visual attributes (e.g., \emph{is black} and \emph{has tail}) and classes~\cite{lampert2009learning}. The correlations allow knowledge transfer from seen to unseen classes, and thus enable ZSL to classify unseen classes based on the shared knowledge~\cite{verma2017simple,li2019leveraging,xu2020attribute,xie2019attentive}. 
Recent ZSL methods have paved the way by adopting extractors or attention for localized attribute knowledge to enhance
semantic learning~\cite{chen2018zero,felix2018multi,xian2018feature,song2018transductive,xie2020region}, but localities may be
biased 
towards high-frequency visual features of seen classes. For example, localities, focusing on capturing highly relevant attributes to classes, may build a biased `short-cut' of individual attributes towards seen classes (e.g.,~\emph{wings} to \emph{Bat}) to maximize the training likelihood. The individual correlations will confuse the prediction of unseen classes with similar features (e.g.,~\emph{Bird}).



To ease biased visual learning, recent ZSL efforts have achieved success in regularizing the learned correlations with attribute groups, i.e., grouping attributes into high-level semantic groups~\cite{atzmon2018probabilistic,jayaraman2014decorrelating,wang2019zero,xu2020attribute,long2017describing}. For example, Xu et al.~\cite{xu2020attribute} refine localities with pre-defined attribute groups (e.g., \emph{hairless} and \emph{furry} belonging to \emph{texture}). The attribute groups can generalize localities from individual correlations to group correlations (e.g., grouping \emph{hairless} and \emph{wings} to \emph{Bat}), which can ease the biased prediction of unseen classes (e.g., grouping \emph{furry} and \emph{wings} to \emph{Bird}). However, when learning complex knowledge, even humans
may need to revisit information 
multiple times to progressively refine and rectify 
the learned knowledge for better understanding~\cite{coelho2016student,clark2000project}. These methods 
directly
use attribute groups to refine localities without any calibration or rectification, which may not be able to learn complex semantics precisely. 


\begin{wrapfigure}{r}{5.1cm}
\vspace{-2mm}
    \centering
    \includegraphics[width=0.35\textwidth]{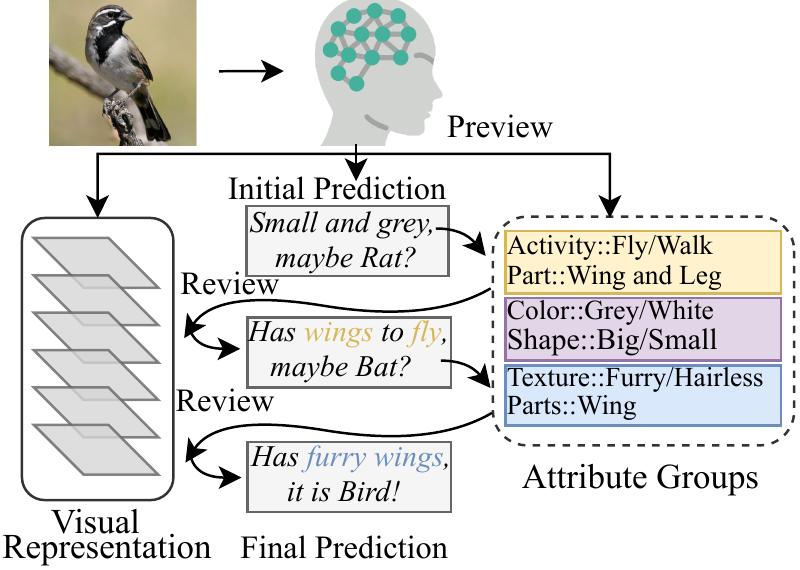}
    \caption{Spiral learning illustration.}
    \label{intro}
    \vspace{-4.3mm}
\end{wrapfigure}
On the other hand, a well-known education paradigm in cognitive theory---\emph{Spiral Curriculum}~\cite{bruner2009process} can teach adults and children to understand complex knowledge by revisits. This curriculum decomposes knowledge into a series of topics. Students first preview overall knowledge and build their views. Then, they gradually select some topics to review and revise 
until they fully understand. Motivated by this, we propose an alternative form of \emph{Spiral Learning} for semantic learning. We take bird classification as an example~(Figure~\ref{intro}). It is challenging to distinguish birds from rats and bats, because they share similar shapes and colors (i.e., purple group in Figure~\ref{intro}). We may preview the image and make an initial wrong prediction---\emph{Rat}. To enhance our view, we can review a combined attribute group of \emph{Activity} and \emph{Part} (i.e., yellow group). By revisiting the visual representation, we can find attributes of \emph{wings to fly} to revise our prediction. Then, the prediction may be confused with \emph{Bat} which can fly. We can further review and find \emph{furry wings} (i.e., blue group), which leads to the final correct prediction---\emph{Bird}. This learning process allows us to spirally accumulate knowledge based on attribute groups, which may ease the difficulty of learning complex semantic correlations.

In this paper, we propose a two-stage Reinforced Self-Revised (RSR) framework to implement spiral learning in an end-to-end manner.
In the preview stage, we design a weakly-supervised self-directed grouping function to automatically group attributes into high-level semantic groups. In the review stage, we propose a reinforced selection module and a revision module to simulate the process that dynamically selects attribute groups to revisit visual information.
Different from 
conventional methods 
that
learn visual localities and directly aggregate them with 
global knowledge, spiral learning aims to learn semantic localities from a global visual representation and thus progressively calibrate the learned knowledge.
We summarize our contributions as follows:
\newline---We propose a novel Reinforced Self-Revised (RSR) framework to decompose conventional learning processes as an incremental spiral learning process. By spirally learning a series of semantic localities based on attribute groups, RSR can dynamically revise the learned correlations and thus ease the difficulty of learning complex correlations. 
\newline---We demonstrate our consistent improvement over state-of-the-art algorithms on four benchmark datasets in both zero-shot and generalized zero-shot settings. To show the extensibility of our framework, we also present an adversarial extension to boost simulation ability.
\newline---We conduct quantitative analysis as well as 
visualization
of 
RSR, which 
indicates that our model can
effectively find significant attributes and combine high-level semantic groups as insightful attribute groups to revise predictions in an incremental way. Moreover, the decision processes are explainable.

\section{Related work}



\textbf{Zero-Shot Learning (ZSL).} The key insight of ZSL is to capture common semantic correlations 
among
both seen and unseen classes. A typical approach is to project visual and/or attribute features to a unified domain, and then apply a compatibility function for classification~\cite{song2018transductive,ye2019progressive,sung2018learning,zhang2017learning,shigeto2015ridge}. Non-end-to-end approaches~\cite{chen2018zero,felix2018multi,xian2018feature} disentangle attributes or generate instances in the embedding space to ease the semantic and visual mismatch.
More closely related to our work, modern end-to-end models~\cite{zhu2019semantic,xie2019attentive,xu2020attribute,yang2020simple} are proposed to extract diverse visual localities 
corresponding to semantics and thus obtain the overall semantic correlation. Zhu et al.~\cite{zhu2019semantic} propose a multi-attention model to obtain multiple discriminative localities under semantic guidance. Xie et al.~\cite{xie2019attentive} further incorporate second-order embeddings to enable stable locality collaboration. 
Some works~\cite{atzmon2018probabilistic,jayaraman2014decorrelating,Liu_2019_ICCV,xu2020attribute,wang2019zero} propose to use attribute groups to enhance  semantic learning. Jayaraman et al.~\cite{jayaraman2014decorrelating} and Atzmon et al.~\cite{atzmon2018probabilistic} group attributes to find joint probabilities of individual attributes for precise prediction. Liu et al.~\cite{Liu_2019_ICCV} use class-specific attribute groups as weights to modify the layer-wise outputs.
Xu et al.~\cite{xu2020attribute} and Long et al.~\cite{long2017describing}
manually group attributes to regularize semantic learning. However, these works 
directly
group attributes for final predictions or rely on extra manual group annotations, which cannot provide multiple complementary attribute groups and may fail to learn groups outside human definitions, e.g., a combined group of multiple manual attribute groups. In our work, we automatically group attributes into diverse groups which can transcend human-defined groups. We learn a series of semantic localities which complement each other during spiral learning.


\textbf{Reinforcement learning in related fields.} Reinforcement learning has been extensively investigated for object detection~\cite{mathe2016reinforcement,pirinen2018deep}, image classification~\cite{wang2020glance,chen2018recurrent}, and few-shot learning~\cite{chu2019spot}. Mathe et al.~\cite{mathe2016reinforcement} and Pirinen et al.~\cite{pirinen2018deep} use reinforcement learning to improve sampling visual regions in an efficient and accurate way. Wang et al.~\cite{wang2020glance} and Chu et al.~\cite{chu2019spot} propose to focus on different visual regions of images and then aggregate regional information to obtain an enhanced overall judgment. Chen et al.~\cite{chen2018recurrent} use a recurrent reinforced module to narrow down the visual space based on the spatial contextual dependency of visual regions. 
These methods aim to use reinforced modules to sample visual regions from visual inputs, and thus narrow down visual space to learn semantic information within regions. Our method is designed to rethink and select the most appropriate attribute groups to guide learning process, which learns semantic information from the visual inputs in a global view governed by the selected attribute groups . 

\begin{figure}
\centering
    \includegraphics[width=0.95\textwidth]{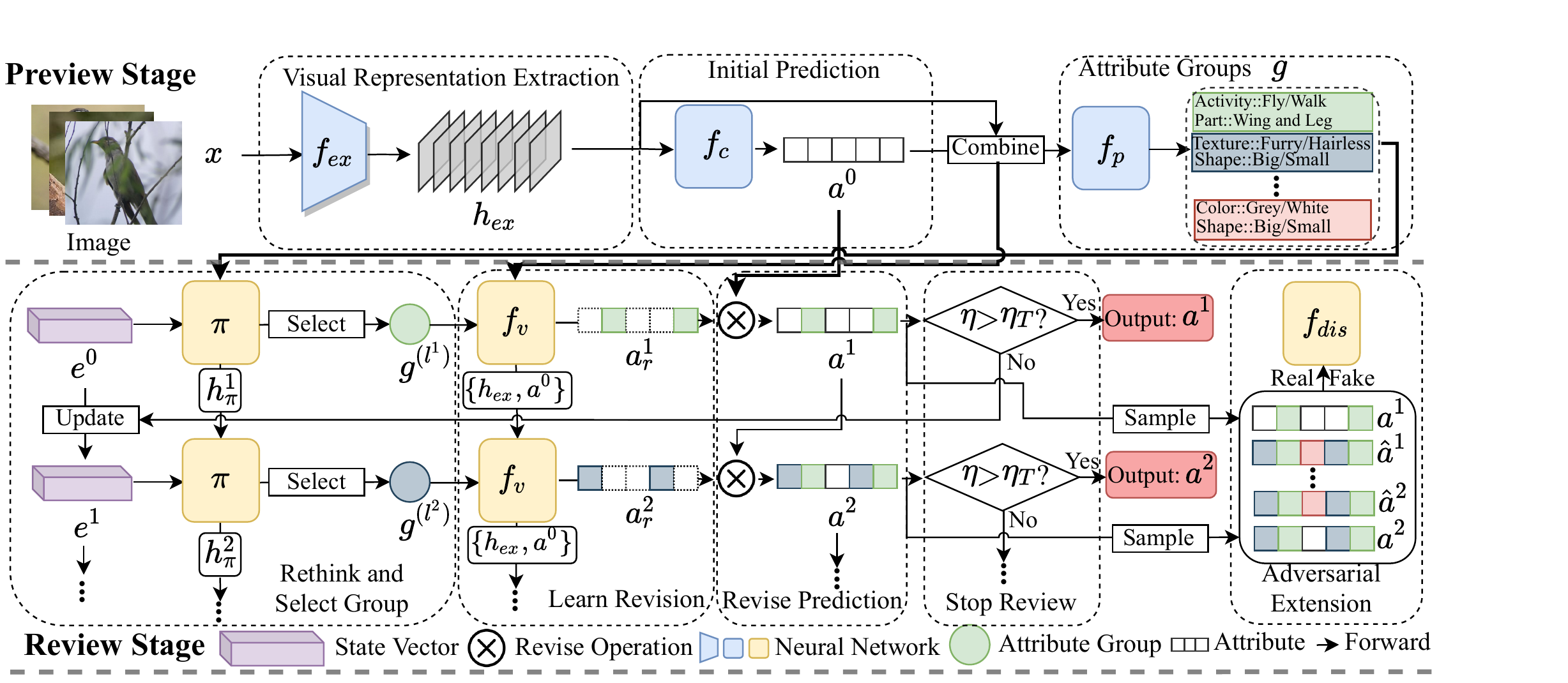}
    \caption{Model overview. Given an instance $x$, RSR spirally learns and revises predictions of the class attribute for $x$. The \emph{preview} stage (upper) extracts a visual representation $h_{\mathit{ex}}$ by extractor $f_{\mathit{ex}}$ and makes initial prediction $a^{0}$ by preview classifier $f_{c}$. A self-directed grouping function $f_{p}$ learns a series of attribute groups as $g$. The \emph{review} stage (lower) spirally revises previous predictions following a sequence of 3R processes. A reinforced module $\pi$ dynamically rethinks and selects $\{(l^{1})^{\mathit{th}},(l^{2})^{\mathit{th}},...\}$ attribute group in $g$. The revision module $f_{v}$ revisits visual representation based on the selected attribute group to learn revisions $\{a_{r}^{1},a_{r}^{2},...\}$. The model uses revisions to revises $a^{0}$ as $\{a^{1},a^{2},...\}$ until the model obtains a confident prediction (i.e., $\eta>\eta_{T}$) or selects all groups. An optional adversarial extension samples and fuses ground-truth attributes $\{\hat{a}^{1},\hat{a}^{2},...\}$ with $\{a^{1},a^{2},...\}$ for an attribute discriminator $f_{\mathit{dis}}$ to enable the model to simulate the prior attribute distribution.}
    \label{model overview}

\end{figure}



\section{Reinforced self-revised network}\label{methodology}
{\bf Problem Formulation.} Given an instance $x\in \mathbb{R}^{H\times W\times C}$ with size $H\times W$ in $C$ channels, each instance has class $y\in \mathbb{N}$ and the corresponding class attribute $\phi(y)=a\in \mathbb{R}^{1\times m}$ with $m$ criteria. Let $X$, $Y$, $A$ be the sets of instances, ground-truth class labels, and pre-defined class attribute vectors, respectively. We define $\mathcal{S}=\{(x, y, a)|x\in{X^{S}},y\in{Y^{S}},a\in{A^{S}}\}$ and $\mathcal{U}=\{(x, y, a)|x\in{X^{U}},y\in{Y^{U}},a\in{A^{U}}\}$ as a training set (i.e.,~seen classes) and a testing set (i.e., unseen classes), respectively. Note that seen classes and unseen classes are disjoint, i.e., $\mathcal{S}\cap \mathcal{U}= \emptyset$, but $A^{S}$ and $A^{U}$ share the same criteria to allow knowledge transfer. Given a test instance $(x,y,a)$, ZSL only predicts unseen instances, i.e., $(x,y,a)\in \mathcal{U}$; Generalized Zero-Shot Learning (GZSL) predicts both seen and unseen instances, i.e., $(x,y,a)\in \mathcal{S}\cup \mathcal{U}$.

Reinforced self-revised network (RSR) consists of two stages: \textbf{Preview} and \textbf{Review}. The preview stage previews instances and learns attribute groups. The review stage consists of \emph{Rethink}, \emph{Revisit}, and \emph{Revise} (3R) processes. The review stage spirally learns semantic localities based on attribute groups to revise decisions in an incremental way. 
A model overview is shown
in Figure~\ref{model overview}.

\subsection{Preview stage}
The preview stage contains three modules: an information extractor $f_{\mathit{ex}}$ for visual representation $h_{\mathit{ex}}$, a preview classifier $f_{c}$ for an initial prediction $a^{0}$, and a self-directed grouping function $f_{p}$ for attribute groups $g$.

\textbf{Visual representation extraction and preview prediction.} Spiral learning incrementally revisits information based on different attribute groups. To reduce the computation cost during revisits, we extract a visual representation $h_{\mathit{ex}}$ for reuse. Considering that different attribute groups may correspond to different parts of inputs, we learn incompletely compressed embeddings by a Convolutional Neural Network (CNN): $f_{\mathit{ex}}(x)\rightarrow h_{\mathit{ex}}$, which can keep the location information. Then, to obtain an initial preview prediction, we use a Fully-Connected Network (FCN) as a preview classifier: $f_{c}(h_{\mathit{ex}})\rightarrow a^{0}$.  We can jointly optimize $f_{c}$ and $f_{\mathit{ex}}$ by a preview cross-entropy loss $\mathcal{L}_{\mathit{PRE}}$:
\begin{equation}\label{non-a-preview}
\min_{f_{\mathit{ex}},f_{c}}\mathcal{L}_{\mathit{PRE}}=-\log\frac{exp(f_{c}(f_{\mathit{ex}}(x))^{T}\phi(y))}{\sum_{\hat{y}\in Y^{S}}exp(f_{c}(f_{\mathit{ex}}(x))^{T}\phi(\hat{y}))}
\end{equation}
where $h_{\mathit{ex}}\in \mathbb{R}^{14\times 14\times 1024}$ denotes visual representation; $f_{c}(f_{\mathit{ex}}(x))\rightarrow a^{0}$ denotes the initial prediction of the preview; $y$ and $\phi(y)$ denote the ground-truth label and true attribute vector of $x$, respectively. 
$\mathcal{L}_{\mathit{PRE}}$ supervises $h_{\mathit{ex}}$ and $a^{0}$ to learn the correct semantic correlations from a global perspective. 
$a^{0}$ can be viewed as an indicator of the learned global semantic correlations.

\textbf{Grouping attributes.} Inspired by \emph{Spiral Curriculum}~\cite{bruner2009process} which splits 
complex concepts into several sub-concepts, we decompose the overall attribute criteria into $k$ diverse sub-groups, i.e., attribute groups, and thus construct different tendencies for semantic learning. We adopt a CNN to project $h_{\mathit{ex}}$ into 2048-dimensional vectors and then combine $h_{\mathit{ex}}$ with $a^{0}$ as inputs for $f_{p}$, indicating the visual information and preview learning state, respectively. We use a FCN as the self-directed grouping function $f_{p}(h_{\mathit{ex}},a^{0})\rightarrow \mathbb{R}_{+}^{k \times m}$ to find $k$ diverse attribute groups as the potential semantic biases that need to be reviewed in $a^{0}$. $f_{p}$ predicts $k$ different weights for each criterion and reshapes the weights followed by a Rectified Linear Unit (ReLU) to deactivate insignificant criteria as $k$ sub-parts of all criteria, i.e., $g\in \mathbb{R}^{km}\rightarrow \mathbb{R}_{+}^{k\times m}$. We summarize $f_{p}$ as follows~\cite{atzmon2018probabilistic}:
\begin{equation}
        g=f_{p}(h_{\mathit{ex}},a^{0})=\mathit{ReLU}(\mathit{Reshape}(\omega_{p}(h_{\mathit{ex}},a^{0})+b_{p}))
\end{equation}
where $\omega_{p}$ and $b_{p}$ denote the weight and bias parameters of $f_{p}$, respectively. 
We optimize $f_{p}$ based on the spiral reviews in the \textbf{review} stage, which enables $f_{p}$ to discover complementary attribute groups. 

\subsection{Review stage}\label{review stage}
The review stage is a sequential process composed of 3R to progressively review and revise $a^{0}$ as $\{a^{1},a^{2},..,a^{t}:t\leq k\}$ without repeated 
attribute groups, which is conducted
by a reinforced selection module $\pi$ and a revision module $f_{v}$. We first introduce review details and then the optimization of $\pi$.

The \textbf{rethink} process progressively selects 
suitable attribute groups to review visual information and eases biased semantic learning in previous predictions. Considering that the selection is progressive and the attribute groups are fixed in the review stage, we design $e^{t}=\{h_{\mathit{ex}},a^{0},a^{t}\}$ as the $t^{\mathit{th}}$ state vector for $\pi$, where $\{h_{\mathit{ex}},a^{0}\}$ can be viewed as the indicator of $g$ and overall biased semantics; $a^{t}$ represents the current state and indicates the solved biases. To enable $\pi$ to progressively `rethink' based on the previous decisions, we use a Gated Recurrent Unit (GRU) in $\pi$ to maintain the previous hidden states $h_{\pi}$. We initialize state vector as $e^{0}=\{h_{\mathit{ex}},a^{0},a^{0}\}$ and hidden state as $h_{\pi}^{0}=\emptyset$. `Rethink' can be summarized: $\pi(e^{t},h^{t})\rightarrow l^{t+1}\in[1,k]$ selecting the $(l^{t+1})^{\mathit{th}}$ group in $g$.

The \textbf{revisit} process learns the revision that extracts semantic information based on attribute groups. Given the selected $(l^{t+1})^{\mathit{th}}$ group in $g$, we first use element-wise multiplication to mask $a^{0}$ with the attribute group to highlight the target semantic locality. Then, we use a FCN as the revision module $f_{v}$ to extract and refine the revision from the visual information $h_{\mathit{ex}}$ by:
\begin{equation}
a_{r}^{t+1} = f_{v}(a^{0},h_{\mathit{ex}},g^{(l^{t+1})})=(\omega_{v}(\{g^{(l^{t+1})}\odot a^{0},h_{\mathit{ex}}\})+b_{v})\odot g^{(l^{t+1})}
\end{equation}
where $a_{r}^{t+1}$ denotes the refined revision by multiplying the selected attribute group $g^{(l^{t+1})}$; $\odot$ denotes 
element-wise multiplication; $\omega_{v}$ and $b_{v}$ are learnable parameters of $f_{v}$.
To enable $a_{r}^{t+1}$ to learn semantic locality which can enhance correlation learning, we use a local cross-entropy loss $\mathcal{L}_{\mathit{LOC}}$:
\begin{equation}
    \mathcal{L}_{\mathit{LOC}}=-\log\frac{\exp((a_{r}^{t+1})^{T}\phi(y))}{\sum_{\hat{y}\in Y^{S}}\exp((a_{r}^{t+1})^{T}\phi(\hat{y}))}
\end{equation}
where $y$ is the ground-truth label; $\phi(y)$ denotes true attribute vector of the input instance. 
The masked revision deactivates insignificant attribute criteria, so only a part of the 
remaining
criteria with large weights in $a_{r}^{t+1}$ will significantly influence $\mathcal{L}_{\mathit{LOC}}$. In other words, $\mathcal{L}_{\mathit{LOC}}$ only optimizes $a^{t+1}_{r}$ to learn generalized semantic locality of the highlighted attribute group. 

The \textbf{revise} process fuses the revision $a^{t+1}_{r}$ with the current prediction to enhance semantic learning. Given the current prediction vector $a^{t}$ and the revision vector $a_{r}^{t+1}$, we propose to revise:
\begin{equation}
    a^{t+1}=\frac{a^{t}}{\left \| a^{t} \right \|}+\beta\frac{a_{r}^{t+1}}{\left \| a_{r}^{t+1} \right \|}
\end{equation}
where $\beta=1/(t+1)$ is an auto-weighted factor to adjust the influence of the revision on the prediction.

We predict labels by finding the most similar class attribute by cosine similarity.
Given $a^{t}$ and $a_{r}^{t+1}$, the similarity of revised $a^{t+1}$ to class $y$ can be calculated as follows:
\begin{equation}
    f_{\mathit{cos}}(a^{t+1},\phi(y))=\frac{1}{\left \| a^{t+1} \right \|}f_{\mathit{cos}}(a^{t},\phi(y))+\frac{\beta}{\left \| a^{t+1} \right \|}f_{\mathit{cos}}(a_{r}^{t+1},\phi(y))
\end{equation}
where $f_{\mathit{cos}}$ is the cosine similarity; $\phi(y)$ is the normalized true attribute. Proof in \textit{Appendix \ref{weighted proof}.}

$f_{\mathit{cos}}(a^{t+1},\phi(y))$ is a weighted similarity 
of $f_{\mathit{cos}}(a^{t},\phi(y))$ and $f_{\mathit{cos}}(a_{r}^{t+1},\phi(y))$, which only propagates the similarity information in revision $a_{r}^{t+1}$ to revise $a^{t}$. In other words, $a^{t+1}$ is the revised prediction of $a^{t}$ by the refined locality $a_{r}^{t+1}$.
To supervise revisions and predictions to complement each other, we use an overall cross-entropy loss $\mathcal{L}_{\mathit{OA}}$ and a joint loss function$\mathcal{L}_{\mathit{JNT}}$:
\begin{equation}
    \mathcal{L}_{\mathit{OA}}=-\log\frac{\exp((a^{t+1})^{T}\phi(y))}{\sum_{\hat{y}\in Y^{S}}\exp((a^{t+1})^{T}\phi(\hat{y}))}
\end{equation}
\begin{equation}
\mathcal{L}_{\mathit{JNT}}=-\log\{\frac{\sum_{i\in[1,m]}[\exp((a^{0})^{T}_{i}\phi(y))][\exp((a^{t+1})^{T}_{i}\phi(y))]}{[\sum_{\hat{y}\in Y^{S}}\exp((a^{0})^{T}\phi(\hat{y}))][\sum_{\hat{y}\in Y^{S}}\exp((a^{t+1})^{T}\phi(\hat{y}))]}\}
\end{equation}
where $(a^{0})_{i}$, $(a^{t+1})_{i}$ denote prediction probabilities of the $i^{\mathit{th}}$ criterion in $a^{0}$ and $a^{t+1}$ by cosine similarity, respectively; $m$ is the attribute criterion dim. $\mathcal{L}_{\mathit{JNT}}$ optimizes the joint probability of $a^{0}$ and $a^{t+1}$. 
We take $\mathcal{L}_{JNT}$ as a regularization term to regularize locality learning to be consistent with the global prediction by modifying the influence of attributes to the optimization based on $a^{0}$.

Then, we can learn the sequence of revised predictions $\{a^{1},a^{2},...,a^{t}\}$ and the corresponding revisions $\{a_{r}^{1},a_{r}^{2},...,a_{r}^{t}\}$. We can summarize the unified review loss function $\mathcal{L}_{\mathit{REV}}$ as follows:
\begin{equation}\label{non-a-review}
    \min_{f_{p},f_{v}} \mathcal{L}_{\mathit{REV}}=\sum_{t}\alpha^{t-1}[\mathcal{L}_{\mathit{LOC}}(a_{r}^{t})+\mathcal{L}_{OA}(a^{t})+\mathcal{L}_{\mathit{JNT}}(a^{0},a^{t})]
\end{equation}
where $\alpha\in (0,1)$ is a pre-defined discount parameter. $\mathcal{L}_{\mathit{REV}}$ enables the review processes to learn refined semantic localities to complement each other in an incremental way, which optimizes attribute groups in $g$ to be constantly linked to diverse and different semantics.

The
\textbf{reinforced selection module} is designed to enhance the review process with better group selection. Thus, $\pi$ aims to improve the accuracy of predicting the ground-truth label $y$ during the review stage. To measure the improvement in accuracy, we define $p(\mathcal{L})_{i}$ as the prediction probability of class $i$ in loss $\mathcal{L}$. The goal of $\pi$ is to improve $\overline{p(\mathcal{L}_{\mathit{REV}})_{y}}$, i.e., the mean correct probability of the terms in $\mathcal{L}_{\mathit{REV}}$. To highlight the revision significance during selection, we define $R_{\mathit{RSR}}=\overline{[p(\mathcal{L}_{\mathit{REV}})_{y}+p(\mathit{UNI})_{y}]}$ as the reward of $\pi$, where $p(\mathit{UNI})_{y}=[p(a^{0})_{y}+\sum_{t}\frac{1}{1+t}p(a_{r}^{t})_{y}]$ is a union probability of the initial prediction and unnormalized revisions to enlarge the probability difference. Then, we optimize $\pi$ by maximizing:
%
\begin{equation}\label{rl loss}
    \max_{\pi} \mathbb{E}[\sum_{t}\gamma^{t-1}R_{\mathit{RSR}}]
\end{equation}
where $\gamma\in (0,1)$ is a pre-defined discount parameter. We implement this optimization by Proximal Policy Optimization (PPO) following \cite{schulman2017proximal}. See \textit{Appendix
\ref{rl optimization}} for more optimization details.

The \textbf{confidence parameter} enables our model to auto-halt the review stage. We define a confidence parameter from the perspective of achieving a solid union prediction $\eta=\max_{i}[p(\mathit{UNI})_{i}]$, which is the highest label probability in the union probability. Given a pre-defined probability threshold $\eta_{T}$, our model early-stops the review stage when obtaining a confident prediction, i.e., $\eta>\eta_{T}$.

\subsection{Adversarial training extension}
Similar to adversarial training~\cite{goodfellow2014generative}, a crucial property of ZSL is to simulate the same semantic correlation in the real semantic distribution. Therefore, adversarial training may enhance the model by using prior attribute distributions to regularize the learned attribute distribution. This section introduces an adversarial version of RSR (A\mbox{-}RSR), which implements adversarial training using the same structure but an additional attribute discriminator $f_{\mathit{dis}}$. 
Considering A\mbox{-}RSR as an extractor, we use $f_{\mathit{dis}}(a^{t})\rightarrow [0,1]$ to distinguish attributes, where `0' denotes the fake attribute from A\mbox{-}RSR and `1' is the true attribute. Then, we optimize an adversarial loss function $\mathcal{L}_{\mathit{A\mbox{-}RSR}}$ to regularize the model to simulate the prior attribute distribution by confusing $f_{\mathit{dis}}$:
\begin{equation}\label{a-rsr-loss}
    \min_{f_{\mathit{ex}},f_{c},f_{p},f_{v}}\max_{f_{\mathit{dis}}}\sum_{t}\alpha^{t-1}\{\mathbb{E}_{\hat{a}^{t}\sim A^{S}}[\log f_{\mathit{dis}}(\hat{a}^{t})]+\mathbb{E}_{a^{t}\sim \mathit{A\mbox{-}RSR}(x)}[\log (1-f_{\mathit{dis}}(a^{t})]+\mathcal{L}_{\mathit{RSR}}(a^{t})\}
\end{equation}
where $\alpha$ is the same discount parameter in RSR; $\hat{a}^{t}\sim A^{S}$ is the ground-truth attribute from the prior attribute distribution of inputs; $a^{t}$ is an attribute from the learned attribute distribution; $\mathcal{L}_{\mathit{RSR}}=\mathcal{L}_{\mathit{PRE}}+\mathcal{L}_{\mathit{REV}}$ is an auxiliary classifier loss to optimize the spiral learning modules.
Then, we split the optimization of $\pi$ into two components to better serve adversarial training. Following Eq.~(\ref{rl loss}), we redesign reward functions as $R_{\mathit{DIS}}=[1-f_{\mathit{dis}}(a^{t})]$, $R_{\mathit{A\mbox{-}RSR}}=[R_{\mathit{RSR}}+f_{\mathit{dis}}(a^{t})]$ for training and confusing $f_{dis}$, respectively. Both $R_{\mathit{DIS}}$ and $R_{\mathit{A\mbox{-}RSR}}$ enable $\pi$ to find the most significant semantic groups which can assist or constrain $f_{\mathit{dis}}$. Thus, we can optimize Eq.~(\ref{rl loss}) based on the new rewards to enable $\pi$ to be consistent with learning goals of Eq.~(\ref{a-rsr-loss}). Note that we do not use $\hat{a}^{t}$ for training $\pi$ to let $\pi$ focus on capturing the significant semantics for A\mbox{-}RSR.

\subsection{Implementation Details}

\textbf{Training strategy.} We propose a 2-stage training procedure to ease the training difficulty: \textbf{Stage-\uppercase\expandafter{\romannumeral1}} optimizes the non-reinforced modules to enable models to provide reliable judgment. We first optimize $f_{\mathit{ex}}$ and $f_{c}$ in the preview stage to extract the reliable visual representation and initial prediction using Eq. (\ref{non-a-preview}). Then, we fix parameters of $f_{\mathit{ex}}$ and $f_{c}$. We use random group selection to optimize $f_{p}$, $f_{v}$, $f_{\mathit{dis}}$ without early-stopping to be capable of handling the general review situations 
following Eq. (\ref{non-a-review}), Eq. (\ref{a-rsr-loss}) for RSR and A\mbox{-}RSR, respectively. \textbf{Stage-\uppercase\expandafter{\romannumeral2}} optimizes the reinforced selection module. We fix the non-reinforced modules and use $\pi$ to select locations with early-stopping, which enables our model to be able to handle the general selection situation and precisely auto-halt in the inference stage. PPO optimizes $\pi$ to maximize reward functions using Eq. (\ref{rl loss}) based on the corresponding rewards for RSR and A\mbox{-}RSR, respectively.

\textbf{Inference strategy.} The model auto-halts the proceeding once $\eta>\eta_{T}$ or 
all groups have been selected,
i.e., $t=k$. Given an arbitrary revised prediction $a^{t}$, we predict labels as follows:
\begin{equation}
     \mathit{ZSL}: \hat{y}=\argmax_{\hat{y} \in Y^{U}} (a^{t})^{T}\phi(\hat{y});\quad \mathit{GZSL}: \hat{y}=\argmax_{\hat{y} \in Y^{U} \cup Y^{S}} [(a^{t})^{T}\phi(\hat{y})-\varepsilon \delta(\hat{y}) ]
 \end{equation}
where $\hat{y}$ denotes the predicted label; $\varepsilon\in[0,1]$ is a calibration factor to fine-tune the model towards unseen classes. $\delta$ is a sign function that returns 1 if $\hat{y}\in Y^{U}$ or 0 otherwise. The second term in GZSL is calibrated stacking~\cite{chao2016empirical}, which is commonly used in end-to-end models~\cite{xu2020attribute,xie2019attentive} to prevent models being largely biased towards seen classes due to the lack of training instances of unseen classes.

\section{Experiments}\label{experiment}
We validate our models on four widely-used benchmark datasets: SUN~\cite{patterson2012sun}, CUB~\cite{welinder2010caltech}, aPY~\cite{farhadi2009describing}, and AWA2~\cite{xian2019zero}. SUN contains 14,340 images of 717 diverse classes with 102 attributes. CUB provides 11,788 images and 312 attributes for 200 types of birds. aPY is a coarse-grained dataset consisting of 15,339 images from 32 classes and 64 attributes. AWA2 comprises 37,322 images of 50 different animals with 85 attributes. The datasets are divided by Proposed Split (PS) to prevent overlapping train/test classes~\cite{xian2019zero}. See more dataset details in \textit{Appendix \ref{detailed_dataset_description}}.

For the comprehensive comparison, we compare our model with 15 state-of-the-art methods including 6 locality-based methods (denoted by $^{*}$). We report these methods in the inductive mode~\cite{xian2019zero} without manual group side information~\cite{xu2020attribute} during training. We use SGD~\cite{bottou2010large} to train our models. We set $\alpha$ as 0.9 and $\gamma$ as 0.99 on four datasets. Grid search (in Section \ref{ablation_study}) is used to set $\eta_{T}$ as 0.4 and $k$ as 5, except $\eta_{T}$=0.7 on aPY for A\mbox{-}RSR. See \textit{Appendix \ref{more_implementation_details}} for more parameter and architecture details.

\subsection{Main results of zero-shot and generalized zero-shot learning}
\begin{table}[t]
\small
\centering
\setlength{\tabcolsep}{1.5pt}
\caption{Main results. $^{*}$ denotes locality-based methods. Base model uses $a^{0}$ as predictions. We measure average per-class accuracy of Top-1 (T1), unseen/seen classes (u/s), and harmonic mean (H).}
\begin{tabular}{l|cccc|ccc|ccc|ccc|ccc}
\hline
\multirow{3}{*}{Method} & \multicolumn{4}{c}{ZSL} & \multicolumn{12}{c}{GZSL}\\\cline{2-17}
&SUN & CUB & aPY & AWA2 & \multicolumn{3}{c}{SUN} & \multicolumn{3}{c}{CUB} & \multicolumn{3}{c}{aPY} & \multicolumn{3}{c}{AWA2} \\\cline{2-17}
&T1 & T1 & T1 & T1 & u & s & H & u & s & H & u & s & H & u & s & H \\
\hline
\hline
\textbf{Non End-to-End}\\
SP-AEN\cite{chen2018zero} & 59.2&55.4&24.1&58.5 & 24.9&\textbf{38.6} & 30.3& 34.7 & 70.6 & 46.6 & 13.7 & 63.4 & 22.6 & 23.0 & 90.9 & 37.1\\
RelationNet\cite{sung2018learning}  & - &55.6&-&64.2 & - & - & - & 38.1 & 61.1 & 47.0 & - & - & -& 30.0 & \textbf{93.4} & 45.3 \\
PSR\cite{annadani2018preserving} & 61.4 & 56.0&38.4&63.8&20.8  &37.2  &26.7&24.6  &54.3   &33.9&13.5  &51.4  &21.4&20.7  &73.8  &32.3\\
$^{*}$PREN\cite{ye2019progressive} & 60.1 & 61.4 & - & 66.6 & 35.4 & 27.2 & 30.8 & 35.2 & 55.8 & 43.1 & - & - & - & 32.4 & 88.6 & 47.4 \\
\cline{1-17}
\textit{Generative Methods}\\
cycle-CLSWGAN\cite{felix2018multi} & 60.0 & 58.4 &- &67.3 &47.9 & 32.4 & 38.7 & 43.8 & 60.6 & 50.8 & - & - & - & 56.0 & 62.8 & 59.2\\
f-CLSWGAN\cite{xian2018feature} & 58.6 & 57.7&-  & 68.2  & 42.6 & 36.6 & 39.4 & 43.7 & 57.7 & 49.7 & - & - & - & 57.9 & 61.4 & 59.6  \\
TVN\cite{zhang2019triple} & 59.3 & 54.9 & 40.9 & 68.8 & 22.2 & 38.3 & 28.1 & 26.5 & 62.3 & 37.2 & 16.1 & \textbf{66.9} & 25.9 & 27.0 & 67.9 & 38.6\\
Zero-VAE-GAN\cite{gao2020zero} & 58.5 & 51.1 & 34.9 & 66.2 & 44.4&30.9&36.5&41.1&48.5&44.4&30.8&37.5&33.8&56.2&71.7&63.0\\
ResNet101-ALE\cite{yucel2020deep}& 57.4 & 54.5 & - & 62.0 & 20.5 & 32.3 & 25.1 & 25.6 & 64.6 & 36.7 & - & - & - & 15.3 & 78.8 & 25.7\\
\hline
\hline
\textbf{End-to-End}\\ 
QFSL\cite{song2018transductive}&56.2&58.8&-&63.5&30.9  &18.5  &23.1 &33.3  &48.1 &39.4&-&-&-&52.1 & 72.8  &60.7\\
$^{*}$SGMA\cite{zhu2019semantic}&- &71.0&-&68.8& - & - & - & 36.7 & 71.3 & 48.5 & - & - & - & 37.6 & 87.1 & 52.5\\
$^{*}$LFGAA\cite{Liu_2019_ICCV}&61.5&67.6&-&68.1& 20.8 & 34.9 & 26.1 & 43.4 & \textbf{79.6} & 56.2 & - & - & - & 50.0 & 90.3 & 64.4\\
$^{*}$AREN\cite{xie2019attentive}&60.6&71.5&39.2&67.9& 40.3 & 32.3 & 35.9 & 63.2 & 69.0 & 66.0 & 30.0 & 47.9 & 36.9 & 54.7 & 79.1 & 64.7\\
$^{*}$SELAR-GMP\cite{yang2020simple} & 58.3 & 65.0 & - & 57.0 & 22.8 & 31.6 & 26.5 & 43.5 & 71.2 & 54.0 & - & - & - & 31.6 & 80.3 & 45.3\\
$^{*}$APN\cite{xu2020attribute}&60.9&71.5&-&66.3& 41.9 & 34.0 & 37.6 & 65.3 & 69.3 & 67.2 & - & - & - & 56.5 & 78.0 & 65.5\\
\hline
\textbf{Ours}\\
Base model (preview) & 60.0 & 68.1& 39.4 &66.9 &43.4&28.6&34.5&61.7&66.4&64.0&34.2&29.7&31.8&\textbf{58.4}&68.0&62.8\\
$^{*}$Random\mbox{-}SR & 63.1 & 71.7 &  43.2 & 68.9 & \textbf{51.0} & 30.8 & 38.4 & 62.8 & 72.9 & 67.5 & 29.9 & 48.1 & 36.9 & 55.4 & 74.8 & 63.7\\
$^{*}$RSR & 64.0 & \textbf{72.1} & 44.2 &\textbf{69.0} & 49.8 & 32.0 & 39.0 & 63.5 & 73.0 & \textbf{68.0} & \textbf{31.8} & 46.1 & 37.6 & 56.0 & 79.1 & \textbf{65.6}\\
$^{*}$Random\mbox{-}ASR & 63.8 & 71.8 & 44.0 & 68.4 & 49.1 & 33.9 & 40.1 & \textbf{65.5} & 69.7 & 67.5 & 30.1 & 49.6 & 38.1 & 56.8 & 74.0 & 64.3 \\
$^{*}$A\mbox{-}RSR & \textbf{64.2} & 72.0 & \textbf{45.4} & 68.4 & 48.0 & 34.9 & \textbf{40.4} & 62.3 & 73.9 & 67.6 & 31.3 & 50.9 & \textbf{38.7} & 55.3 & 76.0 & 64.0\\
\hline
\end{tabular} 
\label{table main}
\end{table}

To validate the effectiveness of the proposed modules, we take the preview stage as the baseline (Base model), which is a fine-tuned ResNet101~\cite{he2016deep} with a classifier. Then, we 
show
the results on the best step during group selections of variants w/o reinforced module or adversarial extension: RSR, Random\mbox{-}SR, Random\mbox{-}ASR, and A\mbox{-}RSR, respectively. See Section \ref{ablation_study} for accuracy per step. Note that Random\mbox{-}ASR and A\mbox{-}RSR use 
Random\mbox{-}SR as the pre-trained backbone. See \textit{Appendix \ref{no_pretrain}} for results without pre-trained backbones. In Table~\ref{table main}, we measure average per-class Top-1 accuracy (T1) in ZSL, Top-1 accuracy on seen/unseen (s/u) classes and their harmonic mean (H) in GZSL. 


In Table~\ref{table main}, compared with the \emph{Base model}, our framework improves the initial prediction by 4.2\%/5.9\% (SUN), 4.0\%/4.0\% (CUB), 6.0\%/6.9\% (aPY), and 2.1\%/2.8\% (AWA2) in ZSL/GZSL.
The progress of Random\mbox{-}SR and Random\mbox{-}ASR prove spiral learning
to be 
effective
at
revising the preview predictions with semantic localities refined by attribute groups. Compared with random selection, RSR and A\mbox{-}RSR further improve random selection by up to 1.4\%/1.9\% in ZSL/GZSL, demonstrating that the self-directed grouping function can provide complementary attribute groups and reinforced selection better 
uses
the group relationships. The adversarial extension slightly impairs the performance of Random\mbox{-}ASR on AWA2. This may be caused by the significant domain shift between seen and unseen classes, which may lead to the model simulating biased seen distributions. Otherwise, the adversarial extension improves performance by up to 1.2\% in both ZSL and GZSL.


Compared with state-of-the-art algorithms, we can observe that the most basic spiral learning model, i.e., Random\mbox{-}SR, can obtain the state-of-the-art performance of locality-based methods, which demonstrates the 
advantage
of learning refined semantic localities based on attribute groups. The proposed reinforced selection module further improves the performance: RSR (e.g.~CUB and AWA2) and A\mbox{-}RSR (e.g.~SUN and aPY) achieve the best performance, which demonstrates the effectiveness of spiral learning in tackling complex correlations. Spiral learning can review different attribute groups and revisit visual information to spirally understand the correlations that cannot be understood with one-time learning. RSR and A\mbox{-}RSR consistently outperform other methods by a large margin, i.e., 2.7\%/2.8\%, 0.6\%/0.8\%, and 4.5\%/1.8\%, on SUN, CUB, and aPY in ZSL/GZSL, respectively. Our models obtain the highest unseen scores and harmonic mean accuracy on four datasets, which demonstrates our effectiveness in learning unbiased localities to revise the learned global semantic correlation. The improvement of performance on CUB and AWA2 is not as significant as that on SUN and aPY, which may be caused by the low accuracy of finding significant attributes and the sparse attribute criterion weights of the learned groups, which will be discussed in Section \ref{exp:self-group}.



\subsection{Attribute group and decision process analysis}\label{exp:self-group}
In this section, we conduct quantitative analysis on the learned attribute groups to demonstrate the effectiveness of the self-directed grouping function, and we visualize the decision process to illustrate the strong explainability of the decision process. We analyze the components of $g$ based on human annotation to discover some cognitive insights. The analysis is conducted from three perspectives: attribute-level, group-level, and decision-level. The criterion number differs in different groups, so we take the maximum 10\% weighted attributes to represent the group tendencies of learning semantics. 

\begin{table*}[h]
\centering
\caption{Attribute-level analysis on $g$. A high sparsity degree indicates weight imbalance.}
\small
\begin{tabular}{|c|cccc|cccc|}
\hline
\multirow{2}{*}{Method} & \multicolumn{4}{c|}{Top-10 shot accuracy} & \multicolumn{4}{c|}{Sparsity degree}\\ \cline{2-9}
& SUN & CUB & aPY & AWA2 & SUN & CUB & aPY & AWA2 \\
\hline
RSR & 0.821 & 0.496	& 1.00 & 0.783 & 1.865 & 2.230 & 2.782 & 3.164\\
A\mbox{-}RSR & 0.816 & 0.487 & 1.00 & 0.685 & 1.921 & 2.166 & 2.272 & 4.467\\
\hline
\end{tabular} 
\label{group_table}
\end{table*}
\textbf{Attribute-level analysis}. We first analyze attribute groups from an attribute perspective, i.e., finding significant attribute criteria and learning a balanced weight distribution. To analyze criterion significance, we view the top-10 largest criteria as the most significant criteria annotated by humans, because we use the feature variance as attribute vectors~\cite{xian2019zero} 
(a
larger value indicates more significant features). Then, given an instance set $X$, we calculate the average \emph{Top-10 shot accuracy} to measure whether $g$ contains the significant attribute criteria by $[\sum_{x\in X}\psi(g_{x})]/|X|$, where $\psi$ denotes a sign function to return 1 if any top-10 largest attribute criterion $a_{i} \in g_{x}$ otherwise 0; $|X|$ denotes the instance number. To measure the balance of weight distribution, we calculate the ratio of the maximum weight to the minimum weight in attribute groups as the \emph{sparsity degree}, i.e., $\max(g)/\min(g)$. When a large sparsity degree exists in attribute groups, it indicates that criterion weights may be imbalanced. We summarize top-10 shot accuracy and sparsity degree in Table \ref{group_table}. We can observe that the self-directed grouping function can effectively find significant criteria from the cognitive judgment of humans, especially aPY. The function can find few significant criteria on CUB, which may lead to the limited improvement of our framework. Criterion weights are balanced on SUB, CUB, and aPY but are slightly imbalanced on AWA2. The high sparsity degree of A\mbox{-}RSR on AWA2 misguides the model to learn a local optimum. See \textit{Appendix \ref{weight_distribution}} for detailed weight distributions.

\begin{wrapfigure}{r}{6.3cm}
\vspace{-4.8mm}
\centering 
\begin{subfigure}{0.42\textwidth}
   \includegraphics[width=\textwidth]{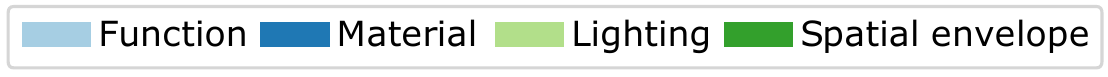}
     \centering
 \end{subfigure}
    
\begin{subfigure}{0.32\textwidth}
   \includegraphics[width=\textwidth]{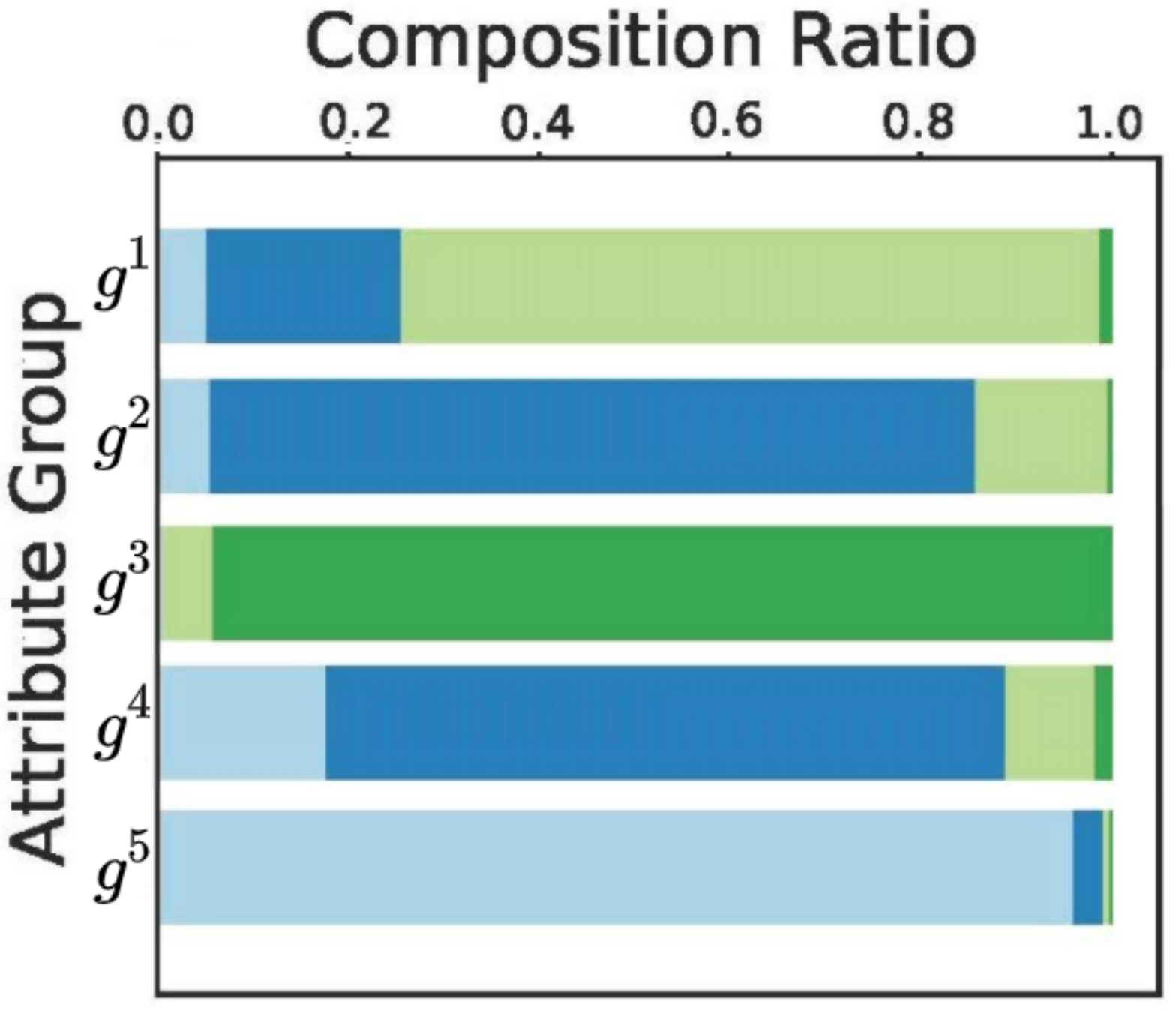}
     \centering
\end{subfigure}
\caption{Semantic analysis of learned attribute groups ($g^{1}\mbox{-}g^{5}$). We annotate the semantic groups (in legend) for attribute criteria in each attribute group based on the human annotations~\cite{patterson2012sun} to reveal the semantic meaning. We plot the relative composition ratios of semantics in each group to illustrate the diverse semantic tendencies of attribute groups.}
\label{group_composition}
\vspace{-6mm}
\end{wrapfigure}

\textbf{Group-level analysis}. To analyze semantic meaning of the learned attribute groups, we refer to human annotations from SUN~\cite{patterson2012sun} which splits attribute criteria into four high-level semantic groups: \emph{function}, \emph{material}, \emph{lighting}, and \emph{spatial envelope} (e.g., \emph{warm} belonging to \emph{lighting}). Note that we do not use manual group annotations during training. We calculate the relative composition ratios of attribute criteria, belonging to different manual semantic groups, to our learned attribute groups as semantic tendencies.
See \textit{Appendix \ref{semantic_analysis_calculation}} for calculation details. In Figure~\ref{group_composition}, we take 5 attribute groups (i.e., $g^{1}\mbox{-}g^{5}$) learned by RSR as examples, and exhibit their average semantic tendencies on SUN. We can observe that the weakly-supervised $f_{p}$ can build five diverse 
semantic tendencies. For example, $g^{5}$ mainly focuses on \emph{function}, while $g^{4}$ combines \emph{material}, \emph{function} and \emph{lighting} as an attribute group, which shows that $g$ can provide insightful attribute groups transcending manual annotations, i.e., combined semantic groups.

\begin{figure*}[h]
    \centering 
\hspace{-1mm}
\begin{subfigure}{0.28\textwidth}
  \includegraphics[width=\textwidth]{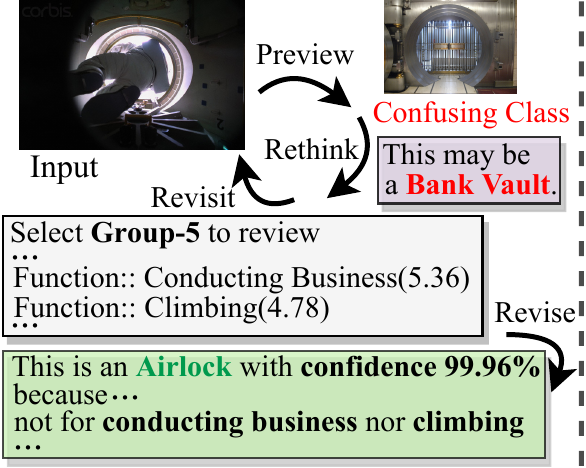}
    \centering
    \caption{Instance-1.}
\end{subfigure}
\begin{subfigure}{0.28\textwidth}
  \includegraphics[width=\textwidth]{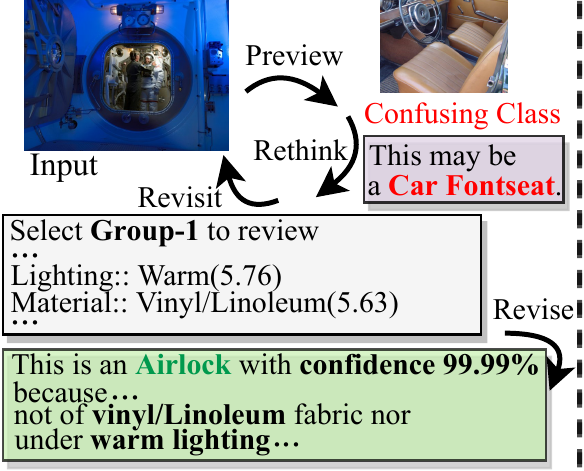}
    \centering
    \caption{Instance-2.}
\end{subfigure}
\begin{subfigure}{0.28\textwidth}
  \includegraphics[width=\textwidth]{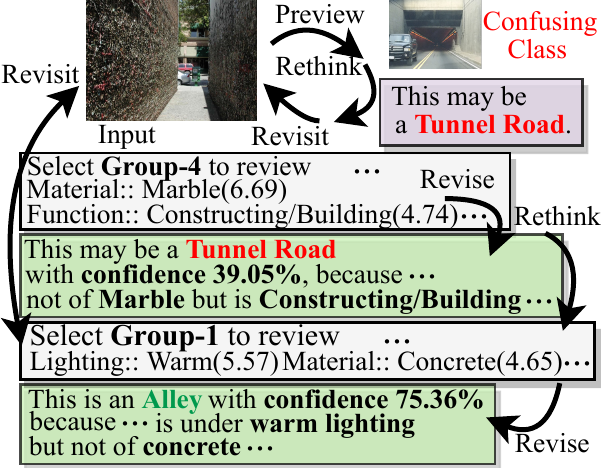}
    \centering
    \caption{Instance-3.}
\end{subfigure}
\caption{Decision process visualization.
Numbers in brackets are specific attribute criterion weights.}
\label{decision_process_visualization}

\end{figure*}

\textbf{Decision-level analysis}. In Figure~\ref{decision_process_visualization}, we visualize three representative instances of \emph{airlock} and \emph{alley}.
From the first two \emph{airlock} instances, we can observe that the reinforced module selects different groups for locality learning due to the different confusing classes. Our model revises the preview prediction \emph{bank vault} and \emph{car fontseat} by learning the business/climbing function, seat material, and lighting, respectively. Comparing instance-2 with instance-3, the self-directed grouping function produces different weights on warm lighting, which indicates that criterion weights are instance-specific. For instance-3 which contains complex correlations, the reinforced module selects suitable groups to progressively distinguish \emph{tunnel road} and \emph{alley} until finding \emph{not of concrete} to make a confident prediction. This dynamic decision process exhibits the strong explainability of our framework.

\subsection{Ablation study}\label{ablation_study}
\begin{figure*}[h]
    \centering 
\begin{subfigure}{0.23\textwidth}
  \includegraphics[width=\textwidth]{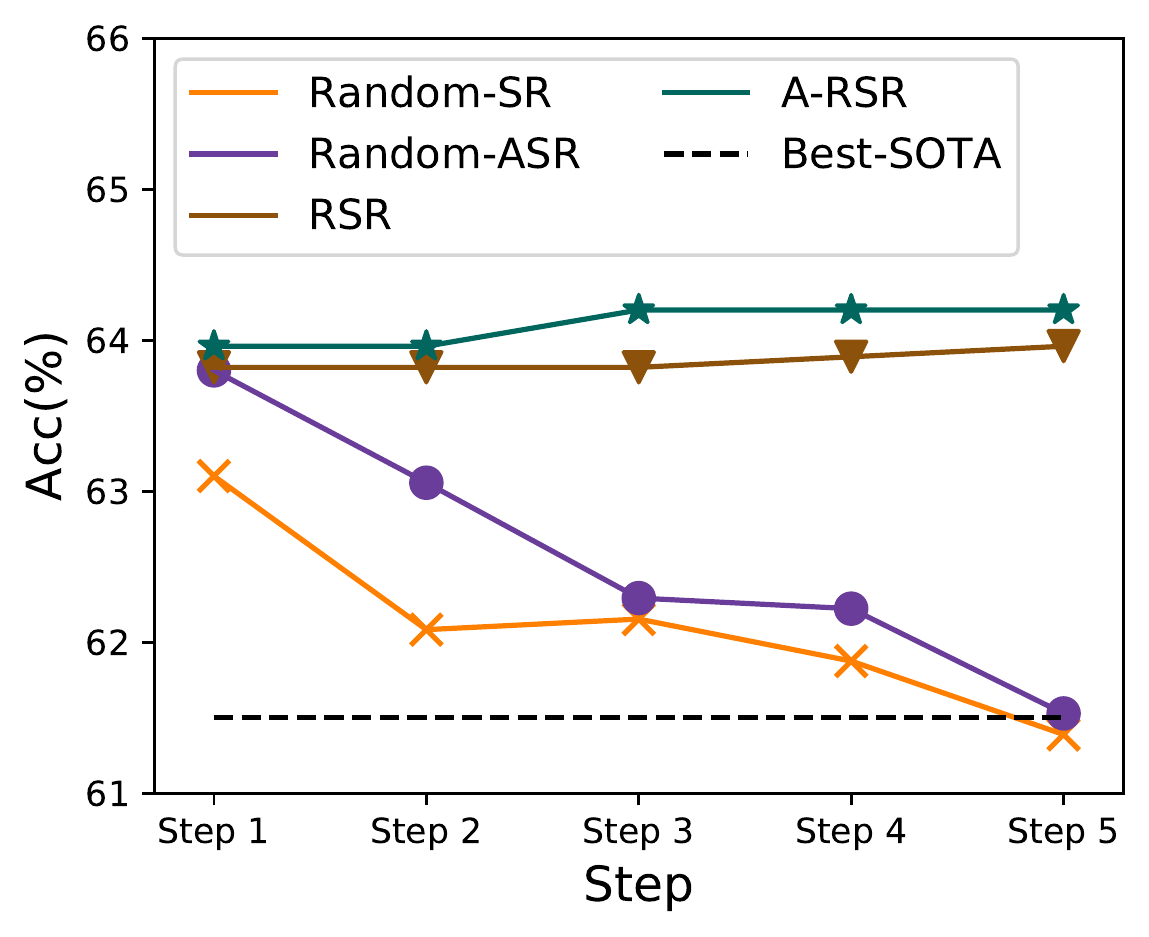}
    \centering
  \caption{SUN.}
\end{subfigure}\hfil 
\begin{subfigure}{0.23\textwidth}
  \includegraphics[width=\textwidth]{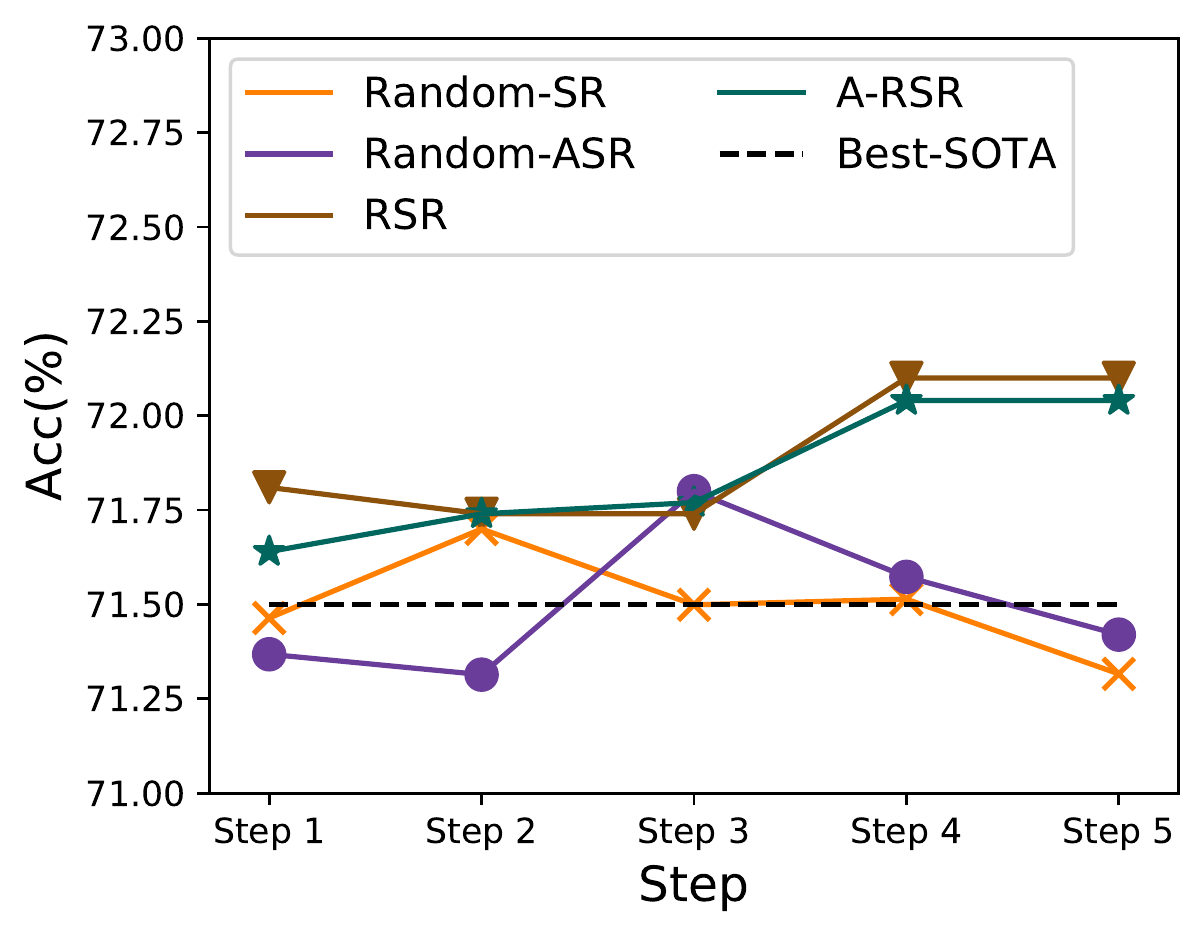}
    \centering
  \caption{CUB.}
\end{subfigure}\hfil 
\begin{subfigure}{0.23\textwidth}
  \includegraphics[width=\textwidth]{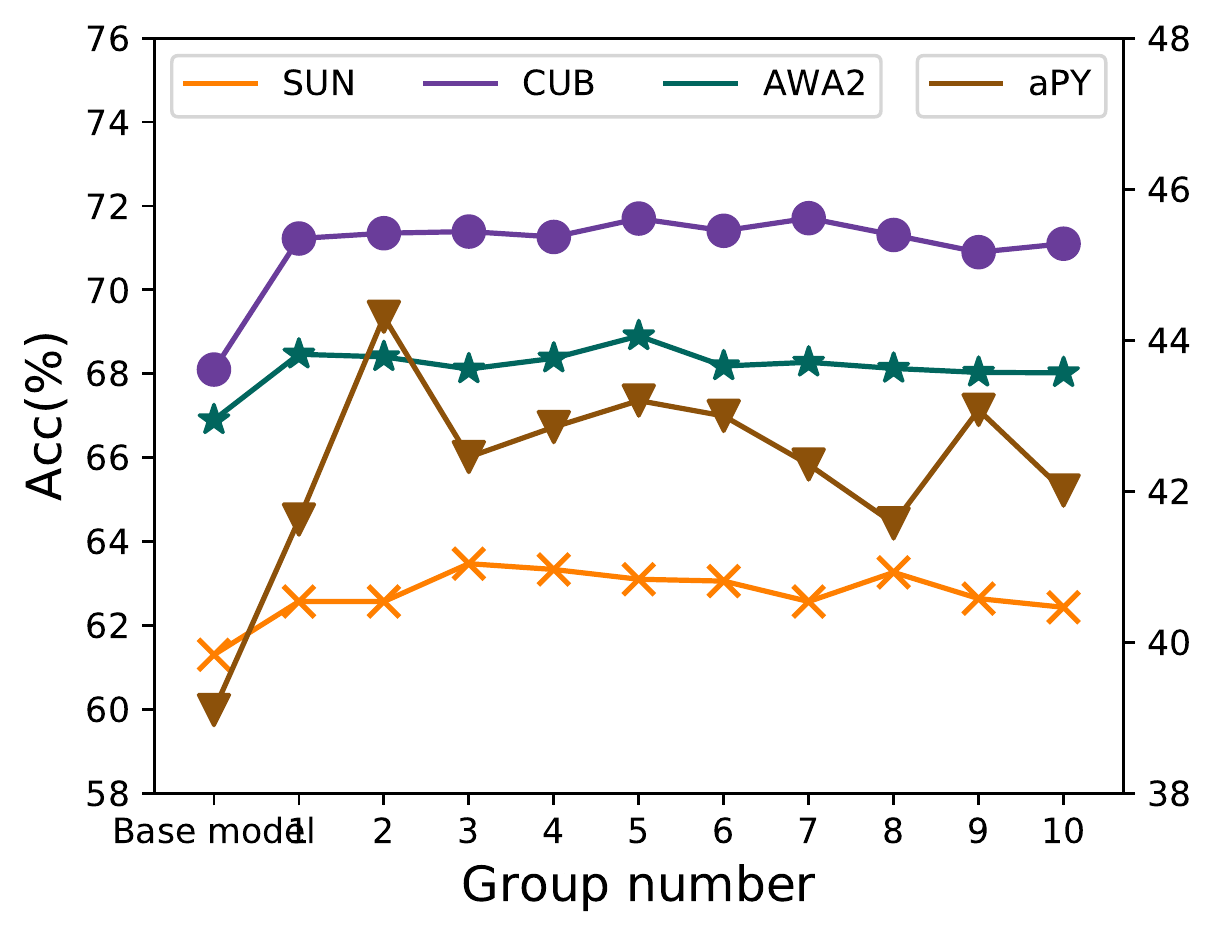}
    \centering
    \caption{Parameter $k$.}
\end{subfigure}
\begin{subfigure}{0.23\textwidth}
  \includegraphics[width=0.95\textwidth]{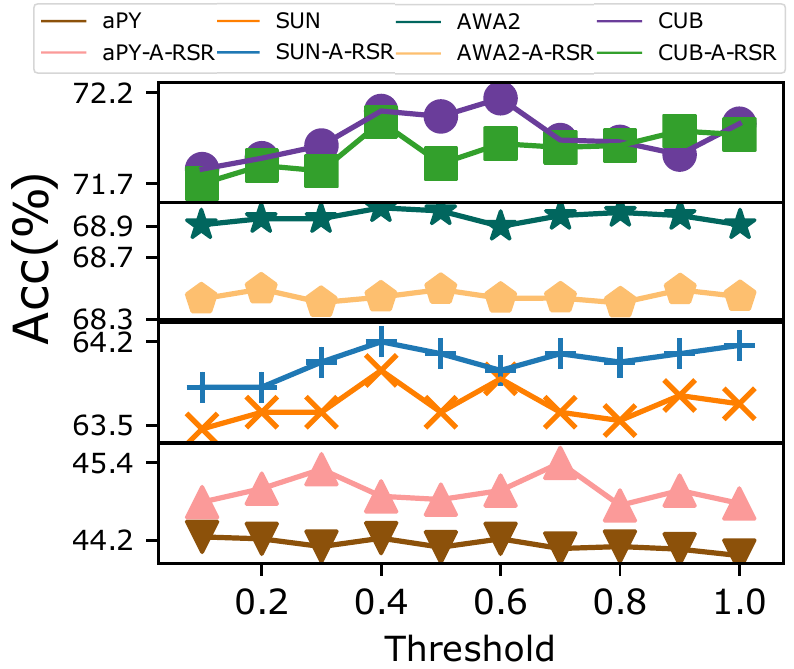}
    \centering
    \caption{Parameter $\eta_{T}$.}
\end{subfigure}
\caption{(a-b) Accuracy curves. (c-d) Hyper-parameter study on $k$ and $\eta_{T}$, respectively.}
\label{ablation study figure}
\end{figure*}
\textbf{Step-accuracy curve.} Figure \ref{ablation study figure}(a)-(b) take SUN and CUB as examples to show accuracy of each step.
Compared with random selection which cannot select suitable attribute groups,
RSR and A\mbox{-}RSR can progressively improve predictions, especially on CUB. This indicates that the reinforced module can utilize complementary attribute groups to guide revisits to visual information and help models incrementally understand some complex correlations which
are difficult to learn
within a single step. 

\textbf{Hyper-parameter study.} 
Figure \ref{ablation study figure}(c) exhibits $k$ analysis of Random\mbox{-}SR which is the initial parameters for variants. $k$ largely influences Random\mbox{-}SR on aPY but less so on other datasets, which indicates that aPY 
can be easy to overfit.
The 
best (on
average)
performance is obtained when $k=5$. Figure \ref{ablation study figure}(d) exhibits $\eta_{T}$ study of RSR and A\mbox{-}RSR. $\eta_{T}$ is sensitive to A\mbox{-}RSR on aPY (i.e., $\pm 0.43$) but stable on other datasets (i.e., $\pm 0.28$).
Overall, our models are stable under different parameters.

\section{Conclusion}
In this work, we present 
a
spiral learning scheme inspired by Spiral Curriculum and propose an end-to-end Reinforced Self-Revised (RSR) framework for ZSL. RSR spirally reviews visual information based on complementary attribute groups to learn the complex correlations that are difficult to capture without revisits. 
The consistent improvement on four benchmark datasets demonstrates the 
advantage
of revisiting semantic localities. We validate the learned attribute groups from quantitative and explainable perspectives, which verifies that the weakly-supervised self-directed grouping function is able to find significant attributes and insightful semantic groups. We also visualize the decision process to illustrate the 
explainability of the spiral learning. The adversarial extensibility of RSR shows 
promise for application in
in other ZSL settings, 
such as generative learning.
\bibliographystyle{plainnat}
\bibliography{bio}

\section*{Checklist}

\begin{enumerate}

\item For all authors...
\begin{enumerate}
  \item Do the main claims made in the abstract and introduction accurately reflect the paper's contributions and scope?
    \answerYes{}
  \item Did you describe the limitations of your work?
    \answerYes{See Section \ref{experiment}.}
  \item Did you discuss any potential negative societal impacts of your work?
    \answerNA{No potential negative societal impacts.}
  \item Have you read the ethics review guidelines and ensured that your paper conforms to them?
    \answerYes{}
\end{enumerate}

\item If you are including theoretical results...
\begin{enumerate}
  \item Did you state the full set of assumptions of all theoretical results?
    \answerYes{See the corresponding parts in Section \ref{methodology}.}
	\item Did you include complete proofs of all theoretical results?
    \answerYes{See \textit{Appendix \ref{weighted proof}}.}
\end{enumerate}

\item If you ran experiments...
\begin{enumerate}
  \item Did you include the code, data, and instructions needed to reproduce the main experimental results (either in the supplemental material or as a URL)?
    \answerYes{We provide code and instructions with pre-trained models to reproduce the results in the supplemental material and URL. The datasets are public, so we provide the downloading link in the supplemental material.}
  \item Did you specify all the training details (e.g., data splits, hyperparameters, how they were chosen)?
    \answerYes{See the first part of Section \ref{experiment} for training details. See Section \ref{ablation_study} \textbf{Hyper-parameter study} for the parameter study.}
	\item Did you report error bars (e.g., with respect to the random seed after running experiments multiple times)?
    \answerNA{We use the fixed seed for all experiments as common ZSL methods. See \textit{Appendix \ref{more_implementation_details}}.}
	\item Did you include the total amount of compute and the type of resources used (e.g., type of GPUs, internal cluster, or cloud provider)?
    \answerYes{See \textit{Appendix \ref{more_implementation_details}}.}
\end{enumerate}

\item If you are using existing assets (e.g., code, data, models) or curating/releasing new assets...
\begin{enumerate}
  \item If your work uses existing assets, did you cite the creators?
    \answerYes{See Section \ref{experiment}.}
  \item Did you mention the license of the assets?
    \answerYes{The assets are commonly-used public datasets and models. See Section \ref{experiment}.}
  \item Did you include any new assets either in the supplemental material or as a URL?
    \answerYes{We provide our codes in the supplemental material and URL.}
  \item Did you discuss whether and how consent was obtained from people whose data you're using/curating?
    \answerNA{Not applicable.}
  \item Did you discuss whether the data you are using/curating contains personally identifiable information or offensive content?
    \answerNA{Not applicable.}
\end{enumerate}

\item If you used crowdsourcing or conducted research with human subjects...
\begin{enumerate}
  \item Did you include the full text of instructions given to participants and screenshots, if applicable?
    \answerNA{Not applicable.}
  \item Did you describe any potential participant risks, with links to Institutional Review Board (IRB) approvals, if applicable?
    \answerNA{Not applicable.}
  \item Did you include the estimated hourly wage paid to participants and the total amount spent on participant compensation?
    \answerNA{Not applicable.}
\end{enumerate}

\end{enumerate}

\newpage
\appendix
\section{More implementation details}\label{more_implementation_details}
We conduct experiments on Python 3.7.9 in Linux 3.10.0 with a GP102 TITANX driven by CUDA 10.0.130 with a fixed seed 272. The neural networks are implemented on Pytorch 1.7.0 and complied by GCC 7.3.0.
\subsection{Dataset description}\label{detailed_dataset_description}
We conduct experiments on four benchmark datasets as follows:

SUN~\cite{patterson2012sun} is a comprehensive dataset of annotated images covering a large variety of environmental scenes, places and the objects. The dataset consists of 14,340 fine-grained images from 717 different classes. Following~\cite{xian2019zero}, the dataset is divided into two parts to prevent overlapping unseen classes from the ImageNet classes: 645 seen classes for training and the rest 72 unseen classes for testing. The attribute is human-annotated and is of 102-dimensions.

AWA2~\cite{xian2019zero} is a subset of AWA dataset, which is an updated version of AWA1. The AWA2 is the only dataset provided with the source images, while the raw images of the AWA1 dataset are not provided. This dataset contains 37,322 images from 50 animal species captured in diverse backgrounds. AWA2 selects 40 classes for training and 10 classes for testing in the ZSL setting. AWA dataset is annotated of binary and continuous attributes, and we take the 85-dimensional continuous attributes for the experiments following~\cite{xian2019zero}, which is more informative.

CUB~\cite{welinder2010caltech} consists of 11,788 images from 200 different bird species. CUB is a fine-grained dataset that some of the birds are visually similar, and even humans can hardly distinguish them. It is challenging that a limited number of instances is provided for each class, which only contains nearly 60 instances. CUB splits classes as 150/50 for train/test in the ZSL setting.

aPY~\cite{farhadi2009describing} comprises of 15,339 images from 32 classes. In the ZSL setting, 20 classes are viewed as seen classes for training, and the remaining 10 classes are used as unseen classes for testing. The dataset is annotated with 64-dimensional attributes. This dataset is very challenging that the classes are very diverse. aPY constitutes two subsets---aYahoo images and Yahoo aPascal images, so there may exist similar objects with different class/attribute semantic correlations.

\subsection{Parameter setting}\label{parameter_set}
For calibrated stacking~\cite{chao2016empirical}, we set $\varepsilon$ to 1.2$\times 10^{-6}$ on SUN, 1.12$\times 10^{-5}$ on CUB, 2.9$\times 10^{3}$ on aPY, and 1.2$\times 10^{-1}$ on AWA2 tuned on a held-out validation set following~\cite{xu2020attribute}, respectively. We use SGD~\cite{bottou2010large} to train our model in an end-to-end manner with momentum of 0.9, weight decay of $10^{-5}$, and learning rate of $10^{-3}$.
\subsection{Network architecture}\label{detailed_network_architecture}
Our framework consists of five modules: visual representation extractor $f_{\mathit{ex}}$, preview classifier $f_{c}$, self-directed grouping function $f_{p}$, revision module $f_{v}$, and reinforced selection module $\pi$. We provide the source code of the testing stage with detailed network architecture codes in \textit{Supplementary files}. 

We let $FCN(m)$ be a fully-connected network layer with output dimension $m$, $Max(7\times 7)$ be a max-pooling layer with kernel size of $7\times 7$, $Adp(1\times 1)$ be an adaptive average pooling layer with output kernel size of $1\times 1$, $Dropout(0.5)$ be a dropout layer with keeping rate of 0.5, $ReLU$ be the rectified linear activation function, and $GRU(1024)$ be a gated recurrent unit with hidden state size of 1024. The parameters that we do not provide are set to the default values.

\textbf{Visual representation extractor $f_{\mathit{ex}}(x)\rightarrow h_{\mathit{ex}}$} is a sub-net of ResNet101 pre-trained on ImageNet~\cite{deng2009imagenet}, which is composed of the blocks before the second-last layer before classifier in the original ResNet101.

\textbf{Preview classifier $f_{c}(h_{\mathit{ex}})\rightarrow a^{0}$} is composed of the second-last layer before classifier in the original ResNet101 pre-trained on ImageNet, a pooling layer, a FCN layer, and a dropout layer. We use $Max(7\times 7)$ on CUB and $Adp(1\times 1)$ on other datasets as the pooling layer. Then, the output of the pooling layer is reshaped as 2048-dimensional vectors. We adopt a 2-layer FCN (i.e., $FCN(1024)-FCN(m)$) on aPY and 1-layer FCN (i.e., $FCN(m)$) on other datasets, where $m$ denotes the attribute dim. Specifically, the 1-layer FCN is without bias term on CUB. The dropout layer keeping rate is set to 0.7 on AWA2 and 0.5 on other datasets.

\textbf{Self-directed grouping function $f_{p}(h_{\mathit{ex}},a^{0})\rightarrow \mathbb{R}_{+}^{k\times m}$} embeds $h_{\mathit{ex}}$ using the second-last module of ResNet101 followed by $Adp(1\times 1)$ and reshapes the output to 2048-dimensional vectors. We concatenate the 2048-dimensional vectors with $a^{0}$ and adopt a 2-layer FCN, i.e., $FCN(1024+m/2)-ReLU-FCN(k*m)-ReLU$ to obtain the attribute groups, where $k$ is the pre-defined group number.

\textbf{Revision module $f_{v}(a^{0},h_{\mathit{ex}},g^{l^{t}})\rightarrow a_{r}^{t}$} adopts the same structure as $f_{p}$ to embed $h_{\mathit{ex}}$ into 2048-dimensional vectors, i.e., the second-last module of ResNet101 followed by $Adp(1\times 1)$. Then, we concatenate the vectors with the masked $g^{(l^{t})}\odot a^{0}$ and feed them to a FCN layer to obtain the revision. We design the FCN layer as $FCN(1024+m/2)-FCN(m)-Dropout$ on aPY and $FCN(m)-Dropout$ on other datasets. Similar to the design of $f_{c}$, we use FCN without bias term on CUB. The keeping rate of dropout layer is set to 0.7 on AWA2 and 0.5 on other datasets.

\textbf{Reinforced module $\pi(e^{t},h_{\pi}^{t})\rightarrow l^{t}$} is a recurrent Actor-critic network, which utilizes a recurrent module to extract information from $e^{t},h_{\pi}^{t}$ to feed actor and critic, respectively. First, we use the concatenated $\{h_{\mathit{ex}},a^{0}\}$ as the component of $e^{t}$, which is composed of the compressed 2048-dimensional $h_{\mathit{ex}}$ and the preview prediction $a^{0}$ from $f_{p}$. We extract information of this part via a FCN layer, i.e., $FCN(1024)$. Then, We use $FCN(m, 256)-ReLU-FCN(512)$ if $m<256$ otherwise $FCN(512)$ to extract information from $a^{t}$. Last, we concatenate the extracted information from $\{h_{\mathit{ex}},a^{0}\}$ and $a^{t}$ with $h_{\pi}^{t}$ to feed $GRU(1536)$ to let the output contain the previous selection information. With the output of GRU, we use $FCN(512)-ReLU-FCN(k)-Softmax$ as actor module to calculate the probability distribution of all the locations, and adopt $FCN(512)-ReLU-FCN(1)$ as critic module to infer the current state score, where $k$ is the pre-defined group number, and the hidden state dim of GRU equals the input dimension.

\section{Proof of Remark 1}\label{weighted proof}
\begin{proof}
Given the current prediction vector $a^{t}$ and the revision vector $a_{r}^{t+1}$, we let $\phi(y)$ be the unit attribute vector of the ground-truth label $y$. Following \cite{dangeti2017statistics}, we can calculate the cosine similarities to $\phi(y)$ for $a^{t}$ and $a_{r}^{t+1}$ as follows:
\begin{equation}
    f_{\mathit{cos}}(a^{t},\phi(y))=\frac{\sum_{i\in [1,m]}a^{t}_{i}\phi(y)_{i}}{\left \| a^{t} \right \| \left \| \phi(y) \right \|}
\end{equation}
\begin{equation}
    f_{\mathit{cos}}(a_{r}^{t+1},\phi(y))=\frac{\sum_{i\in [1,m]}(a_{r}^{t+1})_{i}\phi(y)_{i}}{\left \| a_{r}^{t+1} \right \| \left \| \phi(y) \right \|}
\end{equation}
Similarly, we can easily have:
\begin{equation}
\begin{split}
        f_{\mathit{cos}}(a^{t+1},\phi(y))&=\frac{\sum_{i\in [1,m]}(\frac{1}{\left \| a^{t} \right \|} a^{t}+\frac{\beta}{\left \| a_{r}^{t+1} \right \|}a_{r}^{t+1})_{i}\phi(y)_{i}}{\left \| a^{t+1} \right \| \left \| \phi(y) \right \|}\\
        &=\frac{1}{\left \| a^{t+1} \right \|}f_{\mathit{cos}}(a^{t},\phi(y))+\frac{\beta}{\left \| a^{t+1} \right \|}f_{\mathit{cos}}(a_{r}^{t+1},\phi(y))
\end{split}
\end{equation}
\end{proof}

\section{Proximal Policy Optimization}\label{rl optimization}
Our reinforced selection module is implemented by a recurrent Actor-critic network composed of an actor $\pi$ and a critic $V$, where $V$ aims to estimate the state value~\cite{schulman2017proximal}. During the training process, we sample the location of the selected group $l$ following $l^{t+1}\sim \pi(loc|e^{t},h_{\pi}^{t})$ to optimize the Actor-critic network, where $e^{t}$ denotes the state and $h_{\pi}^{t}$ is the hidden state in the recurrent module for the $t^{\mathit{th}}$ step. We maximize a unified form of reward function for RSR and A\mbox{-}RSR as follows:
\begin{equation}
    \max_{\pi}\mathbb{E}[\sum_{t}\gamma^{t-1}R^{t}]
\end{equation}
where $\gamma=0.99$ is a pre-defined discount parameter and $R_{t}$ denotes the reward function for RSR or A\mbox{-}RSR. According to the work of Schulman et al.~\cite{schulman2017proximal}, the optimization problem can be addressed by a surrogate objective function using stochastic gradient ascent:
\begin{equation}
    \mathcal{L}_{\mathit{CPI}}^{t}=\frac{\pi(l^{t+1}|e^{t},h_{\pi}^{t})}{\pi_{\mathit{old}}(l^{t+1}|e^{t},h_{\pi}^{t})}\hat{f}_{\mathit{ad}}^{t}
\end{equation}
where $\pi_{\mathit{old}}$ and $\pi$ represent the before and after updated policy network, respectively. $\hat{f}_{\mathit{ad}}^{t}$ is the advantages estimated by $V$ as follows:
\begin{equation}
    \hat{f}_{\mathit{ad}}^{t}=-V(e^{t},h_{e}^{t})+\sum_{t\leq i \leq k} \gamma^{i-t}R^{t}
\end{equation}
where $k$ denotes the maximum length of the sampled groups, i.e., the pre-defined group number. The optimization of $\pi$ usually gets trapped in local optimality via some extremely great update steps when directly optimizing the loss function, so we optimize a clipped surrogate objective $\mathcal{L}_{\mathit{CPI}}^{t}$:
\begin{equation}
    \mathcal{L}^{t}_{\mathit{CLIP}}=\min\{\frac{\pi(l^{t+1}|e^{t},h_{\pi}^{t})}{\pi_{\mathit{old}}(l^{t+1}|e^{t},h_{\pi}^{t})}\hat{f}_{\mathit{ad}}^{t},\mathit{CLIP}(\frac{\pi(l^{t+1}|e^{t},h_{\pi}^{t})}{\pi_{\mathit{old}}(l^{t+1}|e^{t},h_{\pi}^{t})})\hat{f}_{\mathit{ad}}^{t}\}
\end{equation}
where $\mathit{CLIP}$ is the operation that clips input to $[0.8, 1.2]$ in our experiments.

Then, we use the following loss function to boost the exploration of the optimal policy network $\pi$ and the precise estimated advantages by $V$:
\begin{equation}
    \max_{\pi,V}\mathbb{E}_{x,t}[L^{t}_{\mathit{CLIP}}-\lambda_{1}\mathit{MSE}(V(e^{t},h_{\pi}^{t},\sum_{t\leq i \leq k} \gamma^{i-t}R^{t})+\lambda_{2}S_{\pi}(e_{t},h_{\pi}^{t})]
\end{equation}
where $\lambda_{1}=0.5$ and $\lambda_{2}=0.01$ are two parameters to smooth the loss function; $\mathit{MSE}$ denotes the mean square error loss function; $S_{\pi}(e^{t},h_{\pi}^{t})$ denotes the entropy bonus for the current Actor-critic network following~\cite{williams1992simple,schulman2017proximal}.

\section{Semantic analysis calculation}\label{semantic_analysis_calculation}
Given an instance $x$, it has $k$ attribute groups $\{g^{1}, g^{2}, ..., g^{k}\}$. From datasets, we have semantic group annotated by humans $\{g_{s}^{1}, g_{s}^{2}, ...\}$. For an attribute group $g^{i}$ and an semantic group annotated by humans $g_{s}^{j}$, we can calculate the semantic ratios of different semantic groups in $g^{i}$ by:
\begin{equation}
    o_{g_{s}^{j}}^{g^{i}}=\frac{|g^{i}\cap g_{s}^{j}|}{|g^{i}|}
\end{equation}
where $o_{g_{s}^{j}}^{g^{i}}$ denotes the ratio of semantic group $g_{s}^{j}$ in attribute group $g^{i}$; $|g^{i}\cap g_{s}^{j}|$ denotes the attribute number of the overlapping attributes in $g^{i}$ and $g_{s}^{j}$; $|g^{i}|$ denotes attribute number in $g^{i}$. Apparently, $\sum_{j}o_{g_{s}^{j}}^{g^{i}}=1$.

To ease the imbalance of criterion numbers in different manual semantic groups, we normalize the ratios to know the relative ratios of semantic groups for each attribute group by:
\begin{equation}
    \mathit{no}_{g_{s}^{j}}^{g^{i}}=\frac{o_{g_{s}^{j}}^{g^{i}}}{\max(\{\left \|o_{g_{s}^{j}}^{g^{i}} \right \|_{2}:i\in[1,k])\}}
\end{equation}
where $\left \|o_{g_{s}^{j}}^{g^{i}} \right \|_{2}$ denotes the L2-norm of $o_{g_{s}^{j}}^{g^{i}}$; $\mathit{no}_{g_{s}^{j}}^{g^{i}}$ denotes the normalized ratio of semantic group $g_{s}^{j}$ in attribute group $g^{i}$.

To better analyze the ratios, we re-scale the ratio scope and let their sum be 1 by Softmax:
\begin{equation}
    \mathit{ro}_{g_{s}^{j}}^{g^{i}}=\frac{\exp(\mathit{no}_{g_{s}^{j}}^{g^{i}})}{\sum_{j}\exp(\mathit{no}_{g_{s}^{j}}^{g^{i}})}
\end{equation}
where $\mathit{ro}_{g_{s}^{j}}^{g^{i}}$ denotes the relative ratio of semantic group $g_{s}^{j}$ in attribute group $g^{i}$; $\sum_{j}\mathit{ro}_{g_{s}^{j}}^{g^{i}}=1$.

Then, let $\mathit{ro}_{g_{s}^{j}}^{g^{i}}(x)$ denotes the ratio of semantic group $g_{s}^{j}$ in attribute group $g^{i}$ for instance $x$. we can know the average relative ratios of semantic groups in attribute groups on each dataset, which can reveal the semantic tendencies:
\begin{equation}
    \mathit{do}_{g_{s}^{j}}^{g^{i}}=\frac{\sum_{x\in X}\mathit{ro}_{g_{s}^{j}}^{g^{i}}(x)}{|X|}
\end{equation}
where $X$ denotes a dataset; $|X|$ denotes the instance number.

We use $\mathit{do}_{g_{s}^{j}}^{g^{i}}$ to portray the semantic tendencies of the learned attribute groups. Only if $g^{i}$ has a high relative ratio for each instance in a dataset, can $\mathit{do}_{g_{s}^{j}}^{g^{i}}$ achieve a high score, which means that $g^{i}$ focuses on learning $g_{s}^{j}$. Otherwise, it does not focus on $g_{s}^{j}$. Therefore, $\mathit{do}_{g_{s}^{j}}^{g^{i}}$ can represent the semantic tendencies of the attribute groups.

\section{More experimental results}
\subsection{Attribute criterion weight distribution analysis}\label{weight_distribution}
\begin{figure*}[htb]
    \centering 
\begin{subfigure}{0.9\textwidth}
  \includegraphics[width=\textwidth]{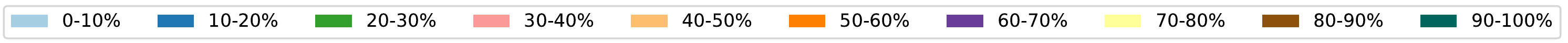}
    \centering
\end{subfigure}
\\
\begin{subfigure}{0.24\textwidth}
  \includegraphics[width=\textwidth]{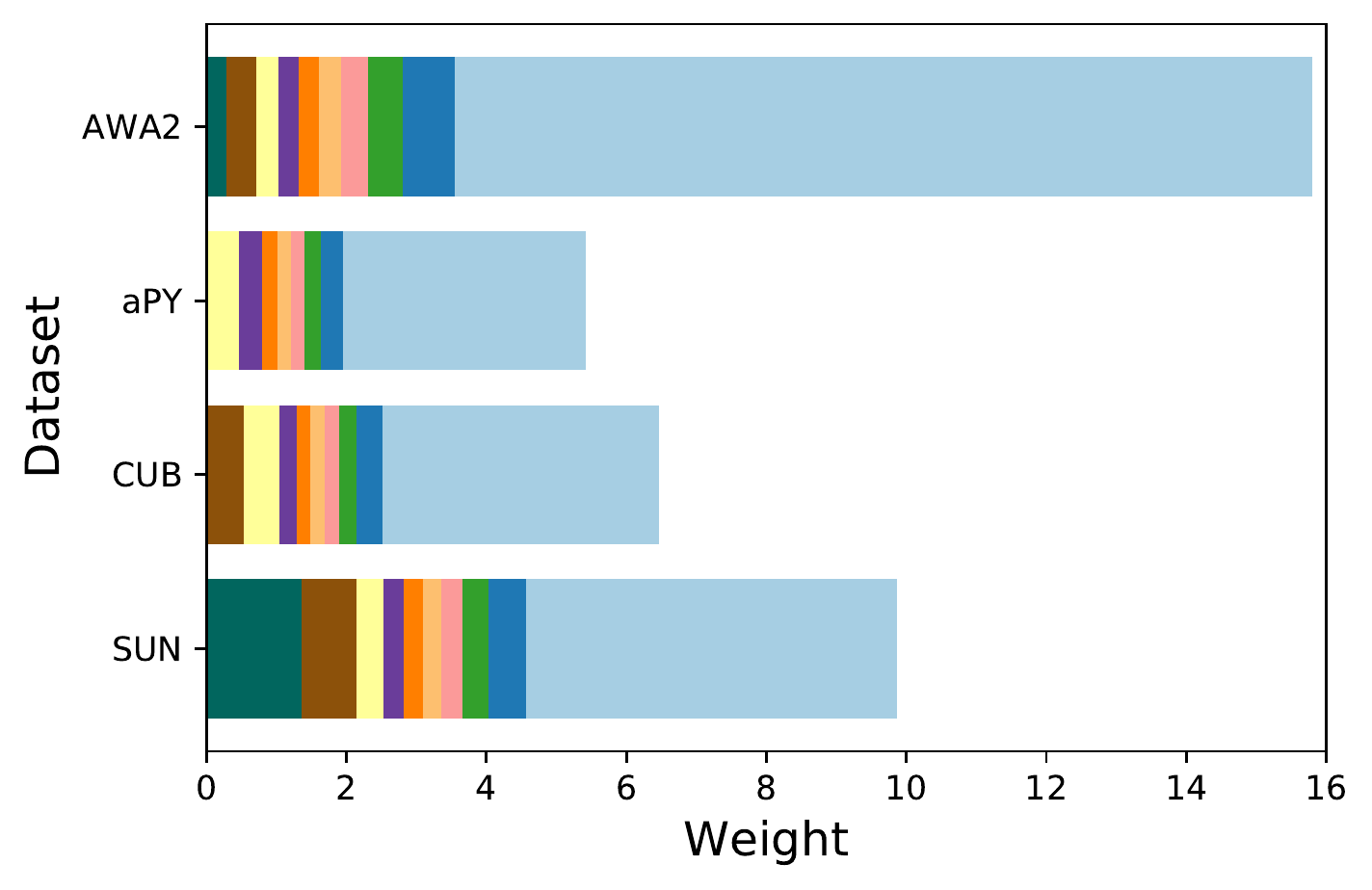}
    \centering
    \caption{RSR training stage.}
\end{subfigure}
\begin{subfigure}{0.24\textwidth}
  \includegraphics[width=\textwidth]{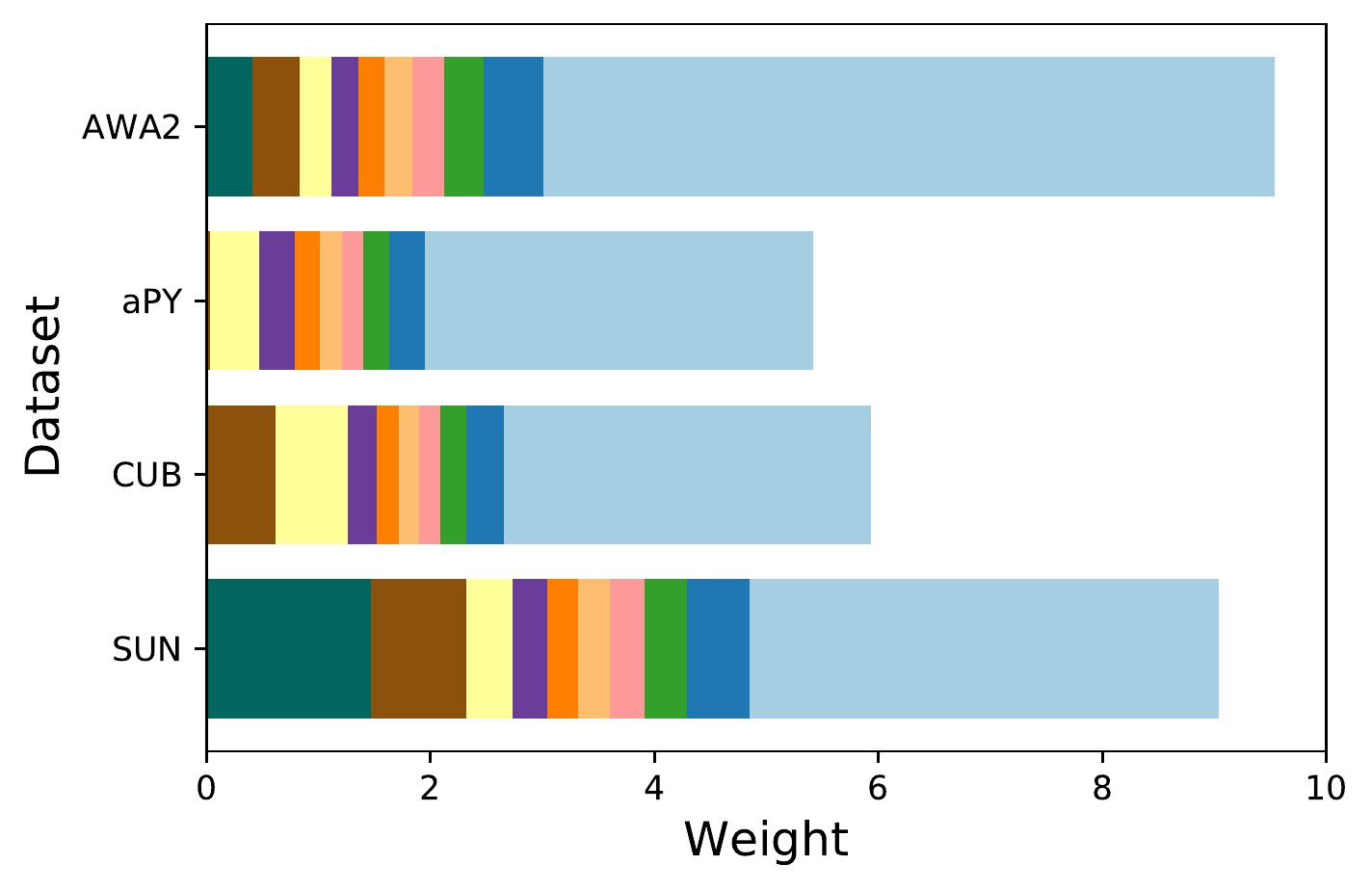}
    \centering
    \caption{RSR testing stage}
\end{subfigure}
\begin{subfigure}{0.24\textwidth}
  \includegraphics[width=\textwidth]{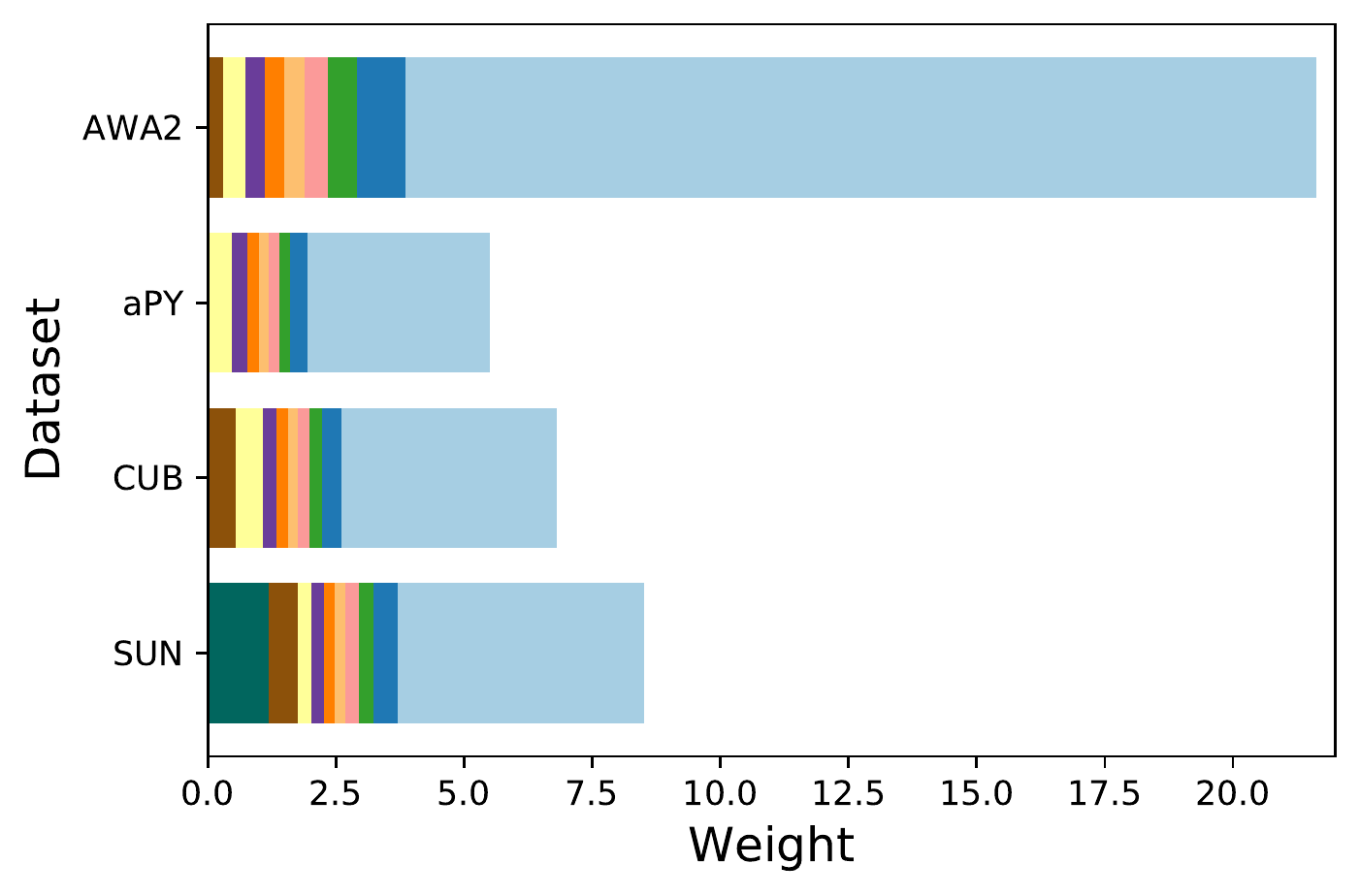}
    \centering
    \caption{A\mbox{-}RSR training stage.}
\end{subfigure}
\begin{subfigure}{0.24\textwidth}
  \includegraphics[width=\textwidth]{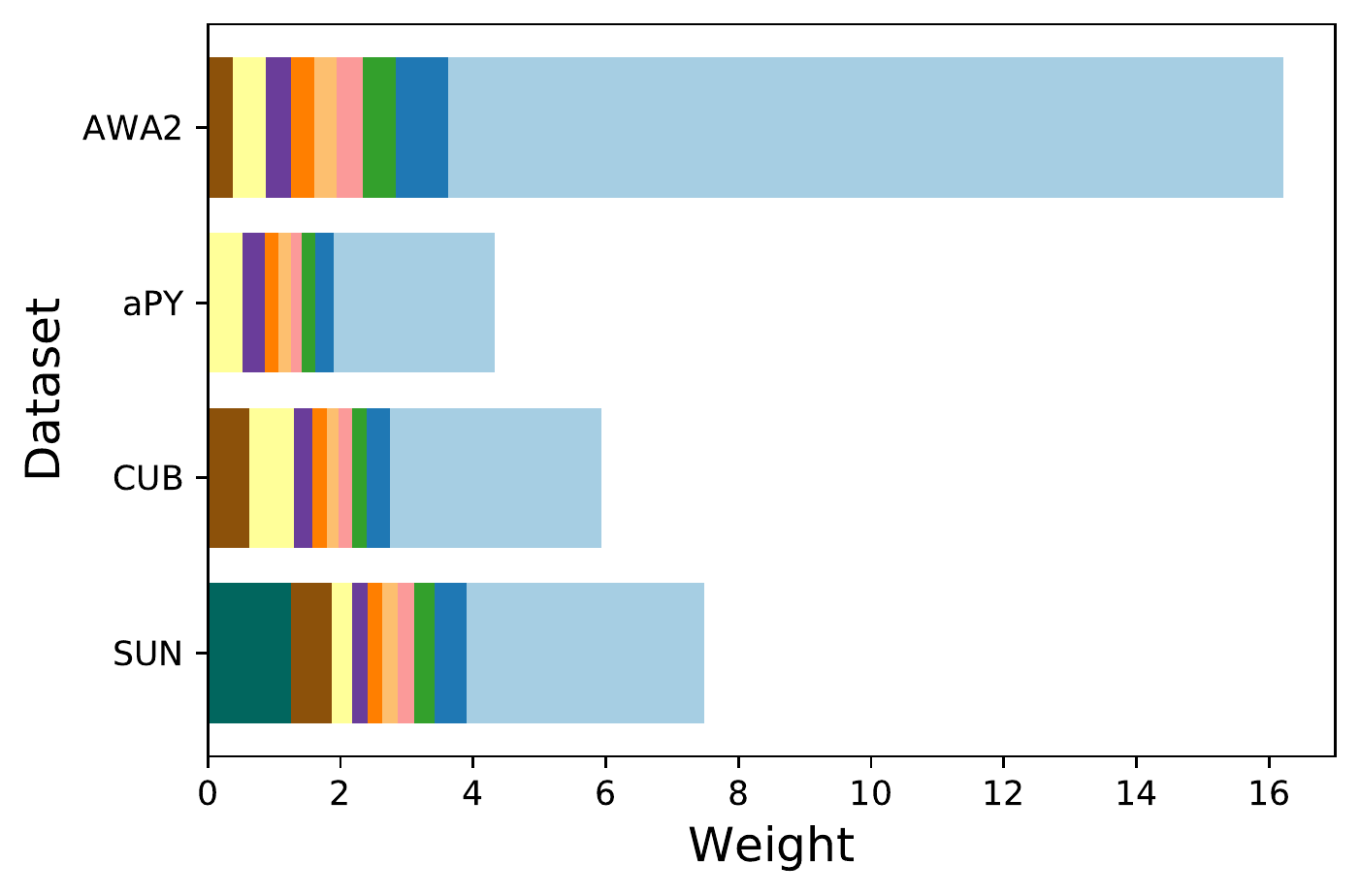}
    \centering
    \caption{A\mbox{-}RSR testing stage.}
\end{subfigure}
\caption{Weight distributions during training and testing of RSR and A\mbox{-}RSR, respectively. The $x$-axis denotes the specific weight value and the $y$-axis denotes different datasets. From left to right in the legend, We sort the weights from low to high and split them into groups with the same sizes of 10\% non-zero attribute criteria.}
\label{weight_distribution_figure}
\end{figure*}

To explore the detailed weight distribution in attribute groups, we sort attribute criterion weights from low to high and split them into groups with the same sizes of 10\% non-zero criteria in Figure~\ref{weight_distribution_figure}. We can observe that most of the weights are small and the model will slightly modify them with the new revisited information. Only the top 10\% criteria take the highest weights in the groups, showing the significant influences during the review stage and can be viewed as the representative semantic tendencies of attribute groups. By comparing the training and testing stages, we can find that the weight distributions of the font 90\% criteria in testing are sparser than those in training, indicating that it is more difficult for the self-directed grouping function to find and concentrate on the significant criteria of the unseen classes. By comparing RSR and A\mbox{-}RSR, we can observe that both models have relatively balanced weight distributions on SUB, which lead to the good performance in ZSL and GZSL experiments. Also, RSR and A\mbox{-}RSR have the most imbalanced weight distributions on AWA2. While RSR shows the better weight distribution in the testing stage than the training distribution, the testing weight distribution of A\mbox{-}RSR tends to be even sparser than the training stage, which leads to the poor performance of A\mbox{-}RSR on AWA2. The imbalanced weight distribution may be caused by the significant domain shift in AWA2, which misguides the adversarial training to simulate the inaccurate semantic correlation.

\subsection{ZSL performance of A\mbox{-}RSR without pre-training}\label{no_pretrain}
\begin{table}[h]
\centering
\caption{ZSL results for non-pre-trained A\mbox{-}RSR.}
\label{table A_RSR_ablation}
\begin{tabular}{lcccc}
\hline
Method & ~~~~~SUN~~~~~& ~~~~~CUB~~~~~ & ~~~~~aPY~~~~~ & ~~~~~AWA2~~~~~  \\ 
\hline
Random\mbox{-}ASR & 62.1&69.6&40.5&68.1\\
A\mbox{-}RSR & 62.2 &71.1&41.1&68.2\\
\hline
\end{tabular} 
\end{table}
In Table~\ref{table A_RSR_ablation}, we show the A\mbox{-}RSR results without using a pre-trained RSR backbone. We can find that the ZSL performance is significantly lower than the results using the pre-trained RSR backbone, decreasing up to 2.0\%, 2.2\%, 4.3\%, 0.2\% on SUN, CUB, aPY, and AWA2. The results indicate that the attribute discriminator is not able to directly regularize the spiral learning. When the base model does not have enough capability of learning an accurate prediction, the adversarial learning may not be able to optimize the base model to obtain overall optimality. Therefore, adversarial training should be a good optional extension of our model to achieve an enhanced spiral learning when the model is well-trained and the domain shift is not significant.


\end{document}